\documentclass[opre,nonblindrev]{informs3}

\OneAndAHalfSpacedXI 

\usepackage{dsfont,mathtools,enumitem,psfrag,bbm,appendix}
\usepackage{algorithm,algorithmic}
\usepackage{times,comment,url}

\def\EE{{\mathbb{E}}}
\def\PP{{\mathbb{P}}}

\def\setR{\mathbb{R}}

\def\setN{\mathbb{N}}

\newcommand{\fall}{\,\forall\,}

\usepackage{natbib}
 \bibpunct[, ]{(}{)}{,}{a}{}{,}%
 \def\bibfont{\small}%
\TheoremsNumberedThrough     
\ECRepeatTheorems

\EquationsNumberedThrough    

\usepackage{color}
\newcommand{\rev}{\color{black}}

\usepackage[colorlinks,
            linkcolor=blue,
            citecolor=blue,
            urlcolor=magenta,
            linktocpage,
            plainpages=false,
            hypertexnames=false,
	    pdfauthor={Siddhartha Banerjee},
            pagebackref=true,
	    pdftex]{hyperref}
\usepackage{bookmark}

\begin{document}

\RUNAUTHOR{Banerjee, Sanghavi and Shakkottai}

\RUNTITLE{Online Collaborative-Filtering on Graphs}

\TITLE{Online Collaborative Filtering on Graphs}

\ARTICLEAUTHORS{%
\AUTHOR{Siddhartha Banerjee}
\AFF{Department of Management Science and Engineering, Stanford University, Stanford, CA 94025\\ \EMAIL{sidb@stanford.edu}} 
\AUTHOR{Sujay Sanghavi, Sanjay Shakkottai}
\AFF{Department of ECE, The University of Texas at Austin, Austin, TX 78705\\ \EMAIL{sanghavi@mail.utexas.edu}, \EMAIL{shakkott@mail.utexas.edu}}
} 

\ABSTRACT{
A common phenomena in modern recommendation systems is the use of feedback from one user to infer the `value' of an item to other users. This results in an exploration vs. exploitation trade-off, in which items of possibly low value have to be presented to users in order to ascertain their value. {\rev Existing approaches to solving this problem focus on the case where the number of items are small, or admit some underlying structure -- it is unclear, however, if good recommendation is possible when dealing with content-rich settings with unstructured content.}

{\rev We consider this problem under a simple natural model, wherein the number of items and the number of item-views are of the same order, and an `access-graph' constrains which user is allowed to see which item. Our main insight is that the presence of the access-graph in fact makes good recommendation possible -- however this requires the exploration policy to be designed to take advantage of the access-graph. Our results demonstrate the importance of `serendipity' in exploration, and how higher graph-expansion translates to a higher quality of recommendations; it also suggests a reason why in some settings, simple policies like Twitter's `Latest-First' policy achieve a good performance.} 

{\rev From a technical perspective, our model presents a way to study exploration-exploitation tradeoffs in settings where the number of `trials' and `strategies' are large (potentially infinite), and more importantly, of the same order. Our algorithms admit competitive-ratio guarantees which hold for the worst-case user, under both finite-population and infinite-horizon settings, and are parametrized in terms of properties of the underlying graph. Conversely, we also demonstrate that improperly-designed policies can be highly sub-optimal, and that in many settings, our results are order-wise optimal.}

}


\maketitle

\section{Introduction}
\label{sec:intro}

The modern internet experience hinges on the ability of content providers to effectively recommend content to users. In such online recommendation settings, user feedback often provides the best guide to the `value' of a piece of content. In content-curation websites  like Digg and Reddit, article recommendation is often done in terms of `popular stories', i.e. content other users found interesting on viewing. In social networks like Twitter and Facebook, each user is shown a (often small) subset of all content generated by her friends/contacts; the selection is based, among other things, on feedback (`likes') from other users. In online advertising, ads that have been shown to a lot of users without much uptake are less likely to work than others with good uptake. 

{\rev Two features are common across these settings: $(i)$ \emph{content-richness}, wherein the amount of available content grows far in excess of what users can consume, and $(ii)$ \emph{unstructured content}, wherein the value of one item of content need not be predictive of the value of other items.} For example, in social networks, every user is both a consumer and also a creator of content, at comparable rates; thus, the available content is of the same order as the total content-views across all users (and far exceeding what a single user can be shown). {\rev Further, content is often unstructured, exhibiting high variability due to periodic trends, one time events, etc. In particular, knowing one piece of content is good need not imply that all content uploaded by the same user is of uniformly high quality. These features make algorithms for recommending relevant content become more critical, but also harder to design.} 

Any system that both recommends items to users, and then leverages their feedback to improve the recommendations, faces an \emph{exploration-exploitation} trade-off: should a user be shown a new item of unknown value, in the hope of benefiting future users? Or should she be shown an item that is already known to give her good value? {\rev This trade-off has been extensively studied in settings wherein the number of items is small, or admit some underlying structure (See Section \ref{sec:relwork} for a discussion of this prior work) -- however it is unclear if these techniques carry over to the applications we describe above.}

Another crucial feature of the settings described above is \emph{the presence of an underlying access graph} between users and items, which constrains what items a user can be presented with. For example, in a social network, users only want to view content uploaded by their friends -- the access graph here is the friendship/follower graph. In content-curation, users may `subscribe' to a set of topics, indicating that they are only interested in content related to these topics. The focus of this paper is to study the effect of such an access graph on recommendation algorithm  design -- {\rev in particular, we suggest that if properly used, the presence of this access-graph may in fact improve the quality of recommendation algorithms.} 

We consider the following stylized model: we are given a bipartite \emph{access graph} between users and items, which specifies which items each user can potentially be shown. Both users and items arrive to the system according to some random process with similar rates -- this captures content-richness. For each visiting user, the algorithm selects a subset of `neighboring items' to present to the user. Each item has an associated value, which can be \emph{arbitrary} -- this captures the unstructured nature of the content. Furthermore, item values are a priori unknown to the algorithm -- to learn them, the algorithm depends on feedback from users. We capture this dependence via the following condition: \emph{for any user, the algorithm can identify the corresponding highest valued items from the set of pre-explored items} -- where an item is said to be pre-explored if the algorithm has presented it to at least one user (or more generally, some finite number of users). The performance of an algorithm is measured in terms of the \emph{competitive-ratio} -- the ratio of the reward that the algorithm earns for an \emph{arbitrary user}, to the best available reward for that user (i.e., the reward earned by a `genie-aided' algorithm, with complete knowledge of the item-values).

In content-rich settings, it is not possible that all items be presented {\em more than} some constant number of times to users. Thus, the act of presenting a popular item to many users may result in other items never being presented to anyone.  In a sense, the critical distinction in content-rich settings is between items for which there are \emph{no ratings}, and those for which there are some -- this is precisely what our model captures. 

Given that the algorithm knows the value of pre-explored items, a sufficient condition for guaranteeing a good per-user competitive-ratio is as follows: \emph{the algorithm should explore items in a manner such that for any user, her most relevant items are explored before she arrives in the system}. It is not hard to see that this is a desirable property, but it may appear too strong a requirement; surprisingly however, we show a milder condition -- the above property holding with a non-vanishing probability -- is in fact achievable in many settings, and using very simple algorithms. On the other hand, competitive performance is by no means guaranteed for all algorithms in this scenario -- we show that certain `natural' algorithms turn out to have vanishing competitive ratio. Furthermore, we derive minimax upper bounds on the competitive ratio which show our results are orderwise optimal in many settings.

{\rev Our results point to three interesting qualitative observations:
\begin{itemize}[nolistsep,noitemsep]
	\item \emph{The role of the access-graph:} We show how the presence of the graph can in fact improve the quality of recommendations; more precisely, we quantify how this improvement depends on certain expansion-like parameters of the graph.
	\item \emph{The importance of serendipity:} Our exploration schemes depend on biasing recommendations towards content from less popular users. Moreover, we show that this is necessary in a very strong sense. 
	\item \emph{The efficacy of simple algorithms:} In particular, our results in the infinite-horizon setting suggest a reason behind the effectiveness of Twitter's `latest-first' recommendation policy.
\end{itemize}
}

Exploration-exploitation trade-offs have been extensively studied in online recommendation literature; in particular, a popular model is the \emph{stochastic bandit model} and its variants. The main assumption in bandit settings is that each item (or arm), upon being displayed, gives an i.i.d reward from some distribution with unknown mean (\cite{Aueretal02a}; see also \cite{BubeckCesaBianchi} for a survey of the field). Algorithms for these settings are closely tied to this assumption -- they focus on detecting suboptimal items via repeated plays, where the number of plays scale with the number of items. This is not feasible in content-rich settings with a very large, possibly infinite, number of arms, and arbitrary values. Indeed using bandit algorithms implies a $0$ competitive-ratio in our setting, which is not surprising as traditional bandit algorithms are designed for a different setting. We discuss this in more detail in Section \ref{sec:relwork}.

We present our results for the case where an item needs to be explored once to know its value; however, our algorithms extend to settings where each item needs a finite number of showings to estimate its value to within a multiplicative factor (as discussed in Section \ref{sec:discussion}). Empirical observations (e.g., by \cite{SzaHub} and \cite{YangLesk}) indicate that this model is reasonable: in large social networks/content-curation sites, \emph{the popularity of an item can be reliably inferred by showing it to a small number of users}. Furthermore, work on \emph{static} recommendation (\cite{keshavan10,jagabathula08}) also provide guarantees for learning item-value from a few ratings under alternate structural assumptions; these observations tie in well with our model.

\subsection{Summary of Our Contributions}

We consider two settings -- a finite-population setting and an infinite-horizon setting. The first is a good model for ad-placement and content-curation, wherein items arrive in \emph{batches}; the infinite-horizon model is more natural for applications like social-network updates, which have continuous arrivals and departures.

\noindent\textbf{Model:} In the finite-population model (Section \ref{ssec:statset}) we assume there is a bipartite access graph between a (fixed) set of $n_U$ users and $n_I$ items -- a user can view an item if and only if she is connected to it. Users arrive in a random order, and are presented with $r$ \emph{item-recommendations}. Each item $i$ has an intrinsic value $V(i)$ -- the total reward earned by a user is the sum of rewards of presented items. The item values are a priori unknown to the algorithm but become known after an item is recommended for the first time. Thus, for any user, the algorithm can always identify the top $r$ pre-explored items.

In the infinite-horizon model (Section \ref{ssec:dynset}), the underlying access-graph $G$ is between a finite set of users $N_u$ and a finite set of \emph{item-classes} $N_C$. The system evolves in time, with user/item arrivals and departures. Each user makes multiple visits to the system, according to an independent Poisson process; similarly, for each item-class, individual items arrive according to an independent Poisson process. Items have arbitrary values, which are a priori unknown; again, we assume that the algorithm can identify the top $r$ pre-explored items for each user. Furthermore, each item is available only for a fixed lifetime. To the best of our knowledge, ours is the first work which provides guarantees for online recommendation under Markovian dynamics but arbitrary item-values.

Our algorithms are as follows: given $r$ slots to present items to an arriving user, we split them between \emph{explore} and \emph{exploit} slots uniformly at random. In the exploit slots, we present the highest-valued pre-explored items (which by our assumption can be identified). For the explore slots, we present \emph{previously unexplored items} -- the crucial ingredient is the \emph{policy for choosing these items}. Our results are as follows:
\begin{enumerate}[noitemsep,nolistsep]
\item  \emph{Exploration via Balanced Partitions}: In the finite setting, we present an algorithm based on picking unexplored items via \emph{balanced semi-matchings} (or balanced item-partitions). We show this achieves a competitive-ratio guarantee of $\Omega(r/d^*(G))$ (Theorem \ref{thm:bser}), where $r$ is the number of recommendations per user, and $d^*(G)$ is the \emph{minimum makespan} of the graph $G$.

\item \emph{Exploration via Inverse-Degree Sampling}: We also present an alternate algorithm that does not use pre-processing, and further \emph{only requires node-degree information}. For each user, the algorithm chooses items for exploration by randomly picking neighborhood items with a probability \emph{inversely proportional to their degree}. This policy achieves a competitive-ratio guarantee of $\Omega(r/Z_{\max}(G))$ (Theorem \ref{thm:ider}), where $Z_{\max}(G)$ is a measure of the \emph{non-regularity} of the graph -- it is greater than the makespan $d^*(G)$, but the two are close when the graph is near-regular. 

\item In the case of regular graphs, both the above algorithms have competitive-ratio guarantees of $\Omega(rn_I/n_U)$. Conversely, in the finite setting, we show that for all graphs, no algorithm can achieve a competitive ratio better than $O(rn_I/n_U)$ (Theorem \ref{thm:convr}).

\item \emph{Exploration via Uniform Latest-Item Sampling}: In the infinite-horizon setting, we propose a competitive algorithm based on discarding items if not explored by their first neighboring user. Each user is presented items drawn  \emph{uniformly and without replacement from the set of latest-items} -- those which have not had the chance to be presented to any prior user. When all arrival processes (of users/items) have rate $1$, we prove that this policy achieves a competitive-ratio of $\Omega(r/Z_{\max}(G))$ (Theorem \ref{thm:ulexp2}).

\item Finally, we show that some intuitive algorithms -- those which always exploit if sufficiently high-valued items are available, or sample nodes uniformly or proportional to degree (or in fact, proportional  to any polynomial function other than inverse-degree) -- have $0$ competitive-ratio.
\end{enumerate}

In both models, our algorithms and results generalize to the setting where an item needs to be viewed by $f$ users to \emph{approximately determine the value} -- within a multiplicative $(1 \pm \delta(f))$ factor for some $\delta(f)\in[0,0.5)$. Further, we do not require for our results that the value be known, but rather, that the top $r$ pre-explored items for a user be \emph{identified} by the algorithm. This allows for various extensions -- in particular, the value can depend on the user identity, i.e., $V(i)$ is replaced by $V(u,i)$ where $i$ corresponds to the item and $u$ to the user identity. We refer to Section~\ref{sec:discussion} for a more detailed discussion.

\section{The Finite-Population Setting}
\label{sec:finite}

We first consider a finite-population setting, where the number of users and items is fixed, and users arrive uniformly at random. This is a good model for certain content-curation problems like news-aggregators (e.g., Google News), where a large number of articles appear together (at the beginning of a day), and expire at the end of the day -- in the meantime, throughout the day, users appear uniformly at random. Furthermore, it also lets us present our main ideas in a more succinct form, avoiding the technical aspects of the infinite-setting while still conveying the main ideas and challenges.

\subsection{System Model}
\label{ssec:statset}

\noindent\textbf{Access Graph:} $G(N_U,N_I,E)$ represents the (given) bipartite access graph between users $N_U$ and items $N_I$ (with $|N_U|=n_U$ and $|N_I|=n_I$). For a user $u\in N_U$, we define its neighborhood as $\mathcal{N}(u):=\{i\in N_I|(u,i)\in E\}$, and degree $d_u=|\mathcal{N}(u)|$; similarly for item $i\in N_I$, we can define $\mathcal{N}(i)$ and $d_i$. Items are always present in the system, while users arrive to the system according to a \emph{uniform random permutation}.

\noindent\textbf{Item Exploration:} Each item has an associated non-negative value $V(i)$, which is a priori unknown; however, presenting it to even one user reveals $V(i)$ exactly. Upon arrival, a user is presented a set of $r$ items from $\mathcal{N}(u)$. We define the $N_I^{expl}$ at any instant to be the set of \emph{pre-explored items}, i.e., which have been presented to at least $1$ user in the past. We assume that $N_I^{expl}=\phi$ at the start; however, all our results hold for any initial $N_I^{expl}$.

\begin{figure}[!ht]
\centering \includegraphics[scale=0.65]{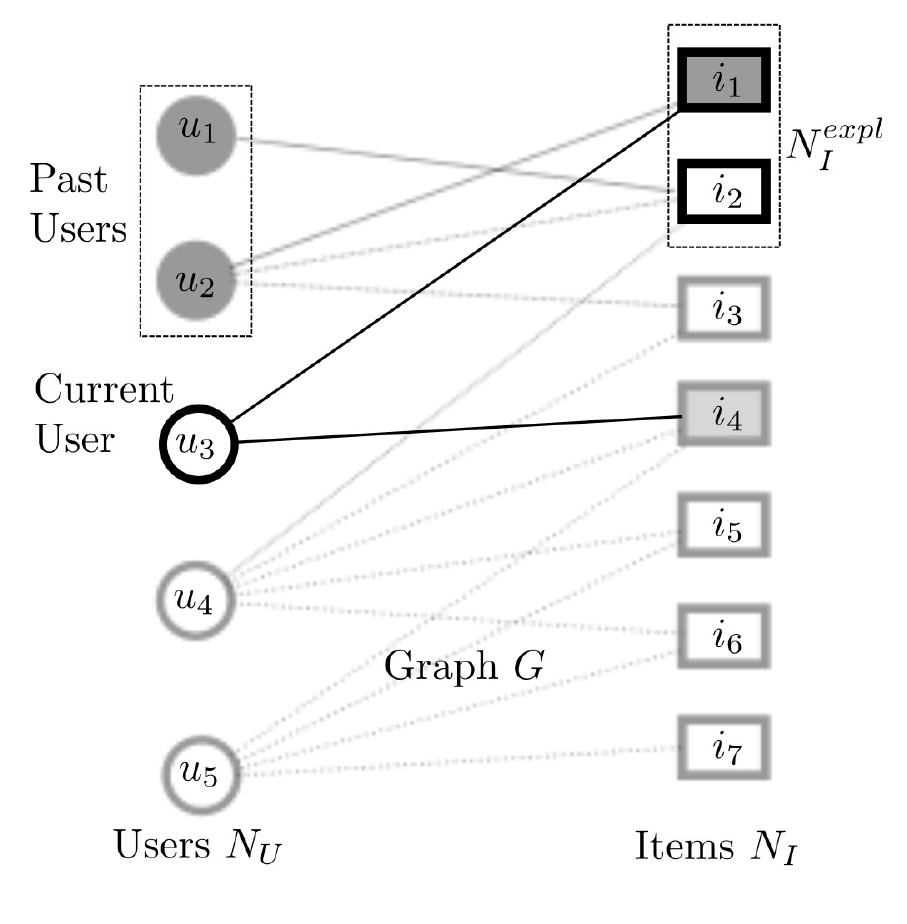}
\caption[Illustration of the finite-population setting]{Illustration of the finite-population setting: Items $i_1$ and $i_4$ (highlighted) have $V(i)=1$ and the rest all have value $0$. Each user is presented $1$ item ($r=1$). Past users $u_1$ and $u_2$ have explored items $i_2$ and $i_1$ respectively. The recommendation algorithm needs to decide which item to recommend to user $u_3$: $i_1$ (exploit) or $i_4$ (explore).}
\label{fig:finite}
\end{figure}

\noindent\textbf{Objective:} For any user $u$, upon arrival, algorithm $\mathcal{A}$ presents $r$ items $\{i^{\mathcal{A}}_1(u)\ldots i^{\mathcal{A}}_r(u)\}$. Thus, for \emph{given item rewards $V$}, the total reward earned by $u$ under algorithm $\mathcal{A}$ is $R_r^{\mathcal{A}}(u)=\sum_{k=1}^rV(i_k^{\mathcal{A}}(u))$. Further, suppose the $r$ highest-valued neighboring items for $u$ be $\{i^*_1(u)\ldots i^*_r(u)\}$ -- then we define the \emph{optimal reward} $R^*_r(u)=\sum_{k=1}^rV(i^*_k(u))$. Finally, we define the \emph{competitive-ratio} $\gamma_r^{\mathcal{A}}(G)$ for algorithm $\mathcal{A}$ (for graph $G$, $r$-recommendations) as:
\begin{equation*}
\gamma^{\mathcal{A}}(G,r)=
\inf_{V\in\setR_+^{n_I}}\inf_{u\in N_U}\frac{\EE\left[R^{\mathcal{A}}_r(u)\right]}{R_r^*(u)}.
\end{equation*}
The expectation here is both over random user-arrivals as well as randomness in algorithm $\mathcal{A}$; however, note that $R^*_r(u)$ is uniquely determined $\forall u$ given $G$ and item-values $V$. The competitive-ratio thus captures a worst case guarantee for individual users and all non-negative item values. {\rev Note that taking an infimum over user-rewards, rather than considering the cumulative reward (i.e., the sum over all users), results in a more stringent objective. However, this is more appropriate in a recommendation setting, as it corresponds to a natural notion of \emph{fairness} -- it is a guarantee on the quality of experience for any user on the platform.}

\subsection{Exploration via Balanced Item-Partitions:}
\label{ssec:bse}

For each user, the algorithm splits the $r$ recommendations between explore and exploit uniformly at random. The exploration step is based on the following pre-processing step: we partition the item-set $N_I$ into $n_U$ sets by associating each item with exactly one of its neighboring users -- we do so in a manner such that the partitions are \emph{balanced}, i.e., we try to minimize the cardinality of the largest set.
\begin{definition}
\label{def:balancepartn}
{\bf (Balanced Partition)} Given graph $G$, a \emph{semi-matching} $M=\{M(u)\}_{u\in N_U}$ is a \emph{partition} of the item-set such that $M(u)\subseteq\mathcal{N}(u)\fall u\in N_U$ (i.e., each set $M(u)$ is a subset of the neighbors of user $u$). Given a semi-matching $M$, we define the load of user $u$ as $d_M(u)=|M(u)|$. Then a \emph{balanced item-partition} $M$ is a solution to the optimization problem:
\begin{equation*}
d^*(G)=\underset{\left\{\mbox{$M$:semi-matching}\right\}}{\mbox{Minimize}}\left[\max_{u\in N_U}d_M(u)\right].
\end{equation*}
\end{definition}
The above problem is known in different communities as the \emph{minimum makespan} problem (\cite{Graham66}), or \emph{optimal semi-matching} problem (\cite{HarveySM}) -- we henceforth refer to $d^*(G)$ as the makespan of graph $G$. Efficient algorithms are known for finding a balanced item-partition, with a complexity of $O(m\sqrt{n}\log n)$ (\cite{FakSM}), where $m=|E|, n=n_U+n_I$. 

{\rev Given a balanced item-partition generation routine, we define the Balanced Partition Exploration Algorithm, or \textbf{BPExp}, which can be summarized as follows: we pre-select a balanced item-partition as an \emph{exploration schedule}; for each arriving user, we independently allocate each `recommendation slot' to be an explore or exploit slot with probability $1/2$; for exploration, we display items picked uniformly at random (without replacement) from the user's items in the balanced item-partition; for exploitation, we display the most valuable available pre-explored items. Formally, the algorithm is given in Algorithm \ref{alg:bser}.}

\begin{algorithm}[!ht]
\renewcommand{\thealgorithm}{1}
\caption{BPExp: Exploration via Balanced Item-Partitions}
\label{alg:bser}
\begin{algorithmic}[1] 
\STATE Generate a \emph{balanced item-partition} $M$ of $G$. Initialize set of explored items $N_I^{expl}=\phi$.
\FOR{arriving user $u\in N_U$}
\STATE 
Choose $R_1(u)\sim Binomial(r,\frac{1}{2})$ slots for \emph{exploration}, and the rest $R_2(u)=r-R_1(u)$ slots for \emph{exploitation}.
\STATE \COMMENT{\textbf{Exploration}}: Choose $R_1(u)$ items from the set $M(u)$  uniformly at random, without replacement.
\STATE \COMMENT{\textbf{Exploitation}}: Recommend the $R_2(u)$ highest-valued items from $\mathcal{N}(u)\cap N_I^{expl}$.
\STATE Update $N_I^{expl}$ by adding the $R_1(u)$ items explored by $u$.
\ENDFOR
\end{algorithmic}
\end{algorithm}

\begin{theorem}
\label{thm:bser}
Given graph $G$, reward-function $V(i)$ and uniformly random user-arrival pattern, using the BPExp algorithm (Algorithm \ref{alg:bser}) we get:
\begin{equation*}
\gamma^{\mbox{\tiny BPExp}}(G,r)\geq\min\left\{\frac{r}{8d^*(G)},\frac{1}{4}\right\}.
\end{equation*} 
\end{theorem}


\noindent\textbf{Remarks:}
\begin{itemize}[nolistsep,noitemsep]
\item[$\bullet$] An immediate corollary of this result is as follows: given \emph{any graph $G$ that contains a perfect matching}, then the competitive-ratio guaranteed by the BPExp algorithm on this graph is $\min\left\{\frac{r}{8},\frac{1}{4}\right\}$. More generally, if $G$ is a bi-regular graph, with $n_I\geq n_U$ (i.e., if all nodes in $N_U$ have the same degree, and similarly all nodes in $N_I$), then $\gamma^{\mbox{\tiny BPExp}}(G,r)\geq \frac{rn_U}{8n_I}$.
\item[$\bullet$] BPExp guarantees a linear scaling with $r$. However, note that we compare the reward earned by BPExp to the optimal reward for $r$ recommendations. In settings where there are $\Omega(r)$ high-valued items, the optimal reward scales linearly with $r$ -- in such cases, BPExp's reward scales \emph{quadratically}. In Section \ref{sec:conv}, we show that linear scaling of $\gamma(G,r)$ with $r$ is in fact the best achievable by any algorithm.
\item[$\bullet$] Consider a graph, where each user is connected to $d^*(G)$ items of degree $1$ -- in this case, it is clear that the best possible competitive-ratio is $\frac{r}{d^*(G)}$. This example is somewhat trivial as it offers no scope for using feedback -- at the other extreme, in Section \ref{sec:conv}, we show that no algorithm can have a better competitive-ratio than $\frac{rn_U}{2n_I}$ in the \emph{complete bipartite graph}, where $d^*(G)=\frac{n_I}{n_U}$. The above theorem shows that on the other hand \emph{$\Omega\left(\frac{r}{d^*(G)}\right)$ is achievable in all graphs}.
\end{itemize} 

\proof{Proof Outline:} Consider the $r=1$ case. The proof now rests on the following observation -- any user $u$ is guaranteed to be presented its corresponding highest-valued item $i^*(u)$ if either: $1.$ the user $u'$ responsible for exploring $i^*(u)$ comes to the system before $u$, or $2.$ $u'$ chose to explore, and explored $i^*(u)$; $u$ chose to exploit. The \emph{former happens with probability $\frac{1}{2}$ due to randomness in user arrivals}. Further, the way we define the BPExp algorithm allows the probability of the latter to be bounded. Combining the two we get the result. We provide the complete proof in Section \ref{sec:proofs}.
\Halmos
\endproof

\subsection{Exploration via Inverse-Degree Sampling:}
\label{ssec:ide}

Although it has a good competitive-ratio, the BPExp has several drawbacks:
\begin{enumerate}[nolistsep,noitemsep]
\item Pre-processing to generate a balanced item-partition is computationally expensive for large graphs.
\item The pre-processing step is inherently \emph{centralized} and requires extensive coordination between the users. This may be infeasible (due to complexity, privacy concerns, etc.).
\item The exploration policy is static. If the underlying graph changes, the item-partition has to be updated.
\end{enumerate}

We now present an alternate approach which overcomes these problems by using a \emph{distributed and dynamic} exploration policy. The main idea is that a user, upon arrival, picks a neighboring item for exploration with a probability \emph{inversely proportional to the degree of the item}. This can be done with minimal local knowledge of the graph  (in fact, the degree information is often publicly available, e.g., followers on Twitter, friends on Facebook/Google+). The resulting competitive-ratio bounds are weaker -- in particular, the makespan $d^*(G)$ is now replaced a quantity $Z_{\max}(G)$, defined as follows:
\begin{equation*}
Z_{\max}(G):=\max_{u\in N_U}Z(u),\quad\mbox{where }Z(u):=\sum_{i\in\mathcal{N}(u)}d_i^{-1}.
\end{equation*}
Note that $Z(u)$ is the normalization in inverse-degree sampling, i.e., when all neighboring items are unexplored, then user $u$ samples item $i$ with $p_{ui}=d_i^{-1}/Z(u)$. To avoid problems with conditioning in the proof, we perform an additional step -- for each user, we partition the neighboring items as follows:
\begin{definition}
\label{def:greedypartn}
{\bf (Greedy Neighborhood Partitioning)} For user $u$, given neighboring-items set $\mathcal{N}(u)$ with degrees $\{d_i\}$, we sort the items in descending order of $d_i^{-1}$ and then generate partition $P_u=\{P_u^1,P_u^2,\ldots,P_u^{r}\}$ by iteratively assigning each item to the set $P_u^k$ with smallest sum-weight.
\end{definition}
{\rev Note that the item-partitioning is performed separately for each user -- it is not a centralized operation.} Given this pre-processing routine, we define the Inverse Degree Exploration Algorithm, or \textbf{IDExp}, as follows:

\begin{algorithm}[!ht]
\caption{IDExp: Exploration via Inverse-Degree Sampling}
\label{alg:ider}
\begin{algorithmic}[1] 
\FOR{arriving user $u\in N_U$}
\STATE Generate item-set partition $P_u=\{P_u^1,P_u^2,\ldots,P_u^{r}\}$ using Greedy Neighborhood Partitioning.
\STATE 
Choose $R_1(u)\sim Binomial(r,\frac{1}{2})$ slots for \emph{exploration}, and the rest $R_2(u)=r-R_1(u)$ slots for \emph{exploitation}.
\STATE \COMMENT{\textbf{Exploration}} Pick $R_1$ sets without replacement from $P_u$, and from each, pick one item $i$ with probability proportional to $d_i^{-1}$.
\STATE \COMMENT{\textbf{Exploitation}}: Recommend the $R_2(u)$ highest-valued items from $\mathcal{N}(u)\cap N_I^{expl}$.
\STATE Update $N_I^{expl}$ by adding the $R_1(u)$ items explored by $u$.
\ENDFOR
\end{algorithmic}
\end{algorithm}


\begin{theorem}
\label{thm:ider}
Given graph $G$, reward-function $V(u,i)$ and uniformly random user-arrivals, the IDExp algorithm (Algorithm \ref{alg:ider}) for recommending $r$ items guarantees:
\begin{equation*}
\gamma^{\mbox{\tiny IDExp}}(G,r) \geq\min\left\{\frac{r}{8eZ_{\max}(G)},\frac{1}{2e}\right\}.
\end{equation*} 
\end{theorem}

\noindent\textbf{Remarks:}
\begin{itemize}[nolistsep,noitemsep]
\item[$\bullet$]  Compared to Theorem \ref{thm:bser}, the above guarantee is weaker by a factor of $\frac{d^*(G)}{Z_{\max}(G)}$ (ignoring constants). In the two extreme cases we considered before (complete bipartite graph, and disjoint item-sets), it is easy to check $Z_{max}(G)$ has the same value as $d^*(G)$; thus we again have that no algorithm can be orderwise \emph{uniformly} better over all graphs.
Further, in case of bi-regular graphs, the two quantities are almost equal (in particular, $Z_{\max}(G)=\frac{n_I}{n_U}$, and $d^*(G)=\lceil\frac{n_I}{n_U}\rceil$). 
\item[$\bullet$] In general, we have $d^*(G)\leq \lfloor Z_{\max}(G)\rfloor$. However there is no $O(1)$-bound in the other direction, and one can construct graphs where $Z_{\max}(G)/d^*(G)$ is $\Omega(n_I)$. This shows that the IDEXP algorithm performs best when the graph is close to regular, but may deteriorate with increasing non-regularity.
\item[$\bullet$] The fact that $Z_{\max}(G)$ is large due to non-regularity can result in the above bound being weak; however, in real-world social network graphs, the performance of the IDEXP algorithm is often much better than the bound. This is because the above bound is for the \emph{worst-case node}; in real-world graphs, removing a few nodes often improves $Z_{\max}(G)$ by a large amount.
\item[$\bullet$] It is somewhat non-intuitive to explore items with a probability \emph{inversely proportional} to its degree -- for example, if an item has the same value for all neighboring users (i.e., $V(u,i)=V(i)$), then not exploring a high-degree item with a high value may seem costly. However, in Section \ref{sec:conv}, we show that inverse-degree randomization is the \emph{only competitive approach} in the following strong sense: any algorithm that explores item $i$ with a probability proportional to $d_i^{-1\pm\epsilon}$ has $0$ competitive-ratio.
\end{itemize}

\proof{Proof Outline:} To see the intuition behind the inverse-degree sampling scheme, note that for any item with degree $d$, each of its neighboring users try to explore it with probability $d^{-1}$ -- thus in a sense, \emph{every item is explored with near-constant probability}. From the point of view of any user $u$, its top item(s) are explored with some constant probability -- further, due to random dynamics, there is a constant probability that the user arrives after the items are explored. We provide the complete proof in Section \ref{ssec:finproof}.
\Halmos
\endproof

\section{The Infinite-Horizon Setting}
\label{sec:infinite}

\subsection{System Model}
\label{ssec:dynset}

We now consider a setting where the {\rev\emph{system evolves in time with user/item arrivals and departures}} -- this is a more natural model for social-network news feeds, and some content-curation sites like Digg/Reddit, where content is posted in a more continuous manner. 

\noindent\textbf{Access graph:} 
We are given an underlying access-graph $G(N_U,N_C,E)$ between \emph{users} $N_U$ and \emph{item-classes} $N_C$ (with $|N_U|=n_U, |N_C|=n_C$). For example, for the problem of generating news-feeds in social networks, the access to user-generated content is restricted by the `follower' graph -- a user can only see updates from people whom she follows. On the other hand, a content-curation website can be viewed as a graph between users and article-topics, with edges incident on a user encoding the personalized set of topics that she is interested in. Each user visits the website periodically to view articles from her topics of interest; correspondingly, for each topic, new articles arrive from time to time. 

\noindent\textbf{User/Item Dynamics:} 
We assume the system evolves in continuous time. Each user generates a series of \emph{visit events} according to a Poisson process of rate $1$. {\rev Equivalently, by the aggregation property of Poisson processes, all user-visits together constitute a marked Poisson process of rate $n_U$ -- each visit 
is denoted by a unique index $s\in\setN_+$ (i.e., a running count of user visits), and has an associated random mark $U(s)$ corresponding to the identity of the visiting user.} A user in each visit is presented $r$ items, chosen from available items in her neighboring item-classes; she accrues rewards from these, provides feedback and leaves instantaneously. 

In parallel, we have an infinite stream of items, where for each item-class, individual \emph{items} arrive according to independent Poisson$(1)$ processes. {\rev As with the users, the item-streams together constitute a marked Poisson process of rate $n_C$ -- each arriving item is denoted by a unique index $i\in\setN_+$, and has three associated parameters: item-class $C(i)$, reward-function $V(i)$, arrival time $T(i)$. Also each item expires after a \emph{fixed} lifetime $\tau$, which we assume is the same for all items.}

\noindent\textbf{Reward Function:} Each item has an arbitrary value -- this can depend on the item class, but not on the specific sample path. {\rev One way to visualize this is as follows -- we allow an adversary to pick a sequence of item-values for each item class -- however the adversary must pick this sequence \emph{before} the user/item arrivals and item recommendation process, and not dynamically as the system evolves (i.e., the adversary is unaware of the sample path of the system when picking the value sequences).} Formally, each item-class $c$ has an associated (infinite) sequence of positive values $V_c$, and the $k^{th}$ item of class $c$ arriving in the system has associated value $V_c(k)$. Note that in any given sample-path $\omega$, the $k^{th}$ item of class $c$ will have associated index $I_{\omega}\geq k$, depending on when it arrives in the system -- by our previous notation, we have $C(I_{\omega})=c$ and $V(I_{\omega})=V_c(k)$.

We say an item-sequence $\mathcal{I}=\{C(i),V(i),T(i),\tau(i)\}_{i\in\setN_+}$ is \emph{valid} if it satisfies the above assumptions. At any time $t$, we define $N_I^{expl}(t)$ to be the set of pre-explored items (i.e., presented during at least one prior visit) currently in the system -- for brevity, we suppress the dependance on $t$. As before, we assume the highest-valued pre-explored items can be identified by the algorithm (see Section~\ref{sec:discussion} for approximate identifiablity from a finite number of user-views).

\begin{figure}[!ht]
\centering \includegraphics[scale=0.80]{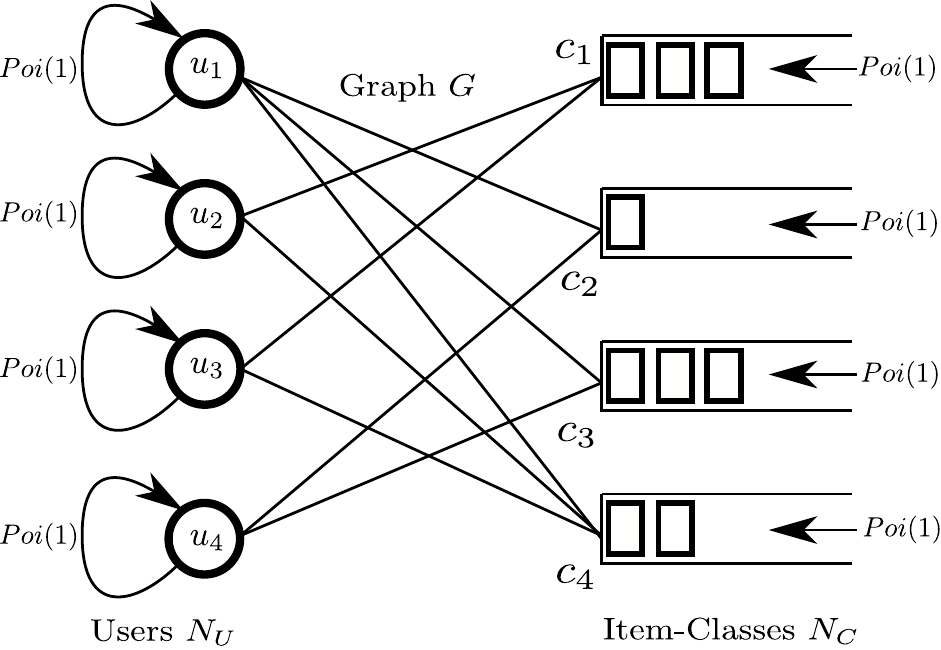}
\caption[Illustration of the infinite-horizon setting]{Illustration of the infinite-horizon setting: There are $4$ users and $4$ item-classes. Users visit according to independent $Poisson(1)$ processes, and similarly items are generated in each class according to an independent $Poisson(1)$ process. Items disappear after a fixed lifetime.}
\label{fig:infinite}
\end{figure}

\noindent\textbf{Objective:} Given valid item-sequence $\mathcal{I}$, and a visit $s$, we define $R^*_r(s)$ as the \emph{optimal offline reward for visit $s$}; note that this is a random variable which depends on which user $U(s)$ corresponds to visit $s$, which items are in the system, etc. Similarly, for a given algorithm $\mathcal{A}$, we can define $R_r^{\mathcal{A}}(s)$. Combining these, we can define the \emph{competitive-ratio} of algorithm $\mathcal{A}$ (given graph $G$, $r$ recommendations) as:
\begin{equation*}
\gamma^{\mathcal{A}}(G,r)=\inf_{\mbox{Valid item-sequence $\mathcal{I}$}}\inf_{\mbox{$s\in\setN_+$}}\EE\left[\frac{R^{\mathcal{A}}_r(s)}{R_r^*(s)}\right].
\end{equation*}

\subsection{Uniform Latest-Item Exploration}
\label{ssec:ulexp}

Given our results for the finite-population setting, a first idea for the infinite-horizon setting would be to apply the IDExp algorithm on the set of available items. This however does not guarantee a competitive ratio, as the number of unexplored items only decreases by at most $r$ after each user-visit. The main idea in designing an exploration policy in this setting is that exploration should be \emph{biased towards more recent items}, while discarding older unexplored items. Let $T_s$ and $T_i$ be the arrival times of visit $s$ and item $i$ respectively. Then, for item $i\in\setN$, we can define its \emph{first neighbor} $S_1(i)$ as the first visit after $T_i$ by a neighboring user (i.e., $S_1(i)=\min\{s|T_s\geq T_i, i\in\mathcal{N}(s)\}$). Correspondingly, for visit $s$, we can define the set of \emph{latest-items} $L(s)$ as the set of available items, for which it is the first neighbor (i.e., $L(s)=\{i|s=S_1(i), T_s<T_i+\tau\}$). Now we have the Uniform Latest-Item Exploration Algorithm (\textbf{ULExp}):
\begin{algorithm}[!ht]
\caption{ULExp: Uniform Latest-Item Exploration}
\label{alg:ulexp2}
\begin{algorithmic}[1] 
\FOR{session $s\in \setN_+$}
\STATE Determine $L(s)$, the set of \emph{latest items}.
\STATE 
Choose $R_1(s)\sim Binomial(r,\frac{1}{2})$ slots for \emph{exploration}, and the rest $R_2(s)=r-R_1(s)$ slots for \emph{exploitation}.
\STATE \COMMENT{\textbf{Exploration}} Pick $R_1$ items from $L(s)$ uniformly at random, and without replacement.
\STATE \COMMENT{\textbf{Exploitation}}: Recommend the $R_2(s)$ highest-valued neighboring items in $N_I^{expl}$.
\STATE Update $N_I^{expl}$ by adding the $R_1(s)$ items explored by $u$.
\ENDFOR
\STATE Remove items from $N_I^{expl}$ when they leave the system.
\end{algorithmic}
\end{algorithm}

Recall in Section \ref{sec:finite}, we defined
$Z_{\max}(G):=\max_{u\in N_U}Z(u),\quad\mbox{where }Z(u):=\sum_{i\in\mathcal{N}(u)}d_i^{-1}$. We now have the following theorem for the competitive-ratio of the ULExp algorithm: 

\begin{theorem}
\label{thm:ulexp2}
Given graph $G$, with both users and items arriving according to independent $Poisson(1)$ processes, using the ULExp Algorithm, we have:
\begin{align*}
\gamma^{\mbox{\tiny ULExp}}(G,r) &\geq\frac{r}{4(5Z_{\max}+2)}.
\end{align*} 
\end{theorem}

\noindent\textbf{Remarks:}
\begin{itemize}[nolistsep,noitemsep]
\item[$\bullet$] We do not need to assume that all the Poisson processes have the same rate -- in fact, in Section \ref{ssec:infproof}, we prove the result for general $\{\lambda_u,\lambda_c\}$. Note that we do not need to know these rates for the algorithm. 
\item[$\bullet$] On the practical side, recommendation via showing the \emph{latest} items, as done on Twitter, can be thought to be a form of uniform latest exploration -- our result suggests that for good recommendation, this should be equally mixed with items which are popular (`trending'). 
\end{itemize}

\proof{Proof Outline:} Using the reversibility of the Poisson processes, we can show that for any visit $s$, the average number of items in its latest-item set is bounded by $Z_{\max}$ -- to see this, note that for visit $s$ and any neighboring item-class $c$, the number of items in $L(s)$ of class $c$ is one less than a $Geometric\left(\frac{d_i}{d_i+1}\right)$ random variable. This suggests that uniform latest-item exploration ensures that any item is explored with high enough probability. 

The technical difficulty arises in the fact that for any visit, we want such a guarantee for the corresponding highest-valued item for that visit -- this item is selected based on the sample-path and the sequence of rewards, which is \emph{arbitrary}. Note that the rewards can affect which item is the highest-valued -- for example, it the sequence of item rewards is strictly decreasing, then the most valuable item is the oldest available item. Thus we can not argue that the probability of the highest-valued item being explored is the same as that of a typical item. We present a more refined counting argument that accounts for this conditioning -- the complete proof is given in Section \ref{sec:proofs}.
\Halmos
\endproof

\section{Converse Results}
\label{sec:conv}

We now present some converse results, which put in perspective the performance of our algorithms. We present two types of results -- upper bounds on the competitive-ratio \emph{over all possible online algorithms}, and negative results ($0$ competitive-ratio) for \emph{specific algorithms}. All results in this sections are for the finite-population setting.

\noindent\textbf{Upper Bounds:} For our upper bounds, we consider a complete bipartite access-graph, and \emph{binary rewards} -- wherein each item has a value $V(i)\in\{0,1\}$. In this setting, we show that \emph{no algorithm can achieve a competitive-ratio better than $n_U/2n_I$}. Note that for these graphs $Z_{\max}(G)=\frac{n_U}{n_I}$, and $d^*(G)=\lceil\frac{n_U}{n_I}\rceil$. 

\begin{theorem}
\label{thm:conv1}
Given any $\epsilon>0$ and $n_U$, there exists a sufficiently large $n_I$ such that for a $n_U\times n_I$ complete bipartite access-graph, no algorithm can achieve $\gamma(G,1)>\frac{n_U}{2n_I}+\epsilon$.
\end{theorem}

\noindent Moreover, for $r$ item recommendations, \emph{no algorithm can achieve better than linear scaling in $r$}:

\begin{theorem}
\label{thm:convr}
Given any $\epsilon>0$, $n_U$ and $r$, there exists a sufficiently large $n_I$ such that for a $n_U\times n_I$ complete bipartite access-graph, no algorithm which is allowed to show at most $r$ recommendations per user can achieve $\gamma(G,r)>\frac{rn_U}{2n_I}+\epsilon$.
\end{theorem}

Together, these results show that \emph{our competitive-ratio bounds are the best possible up to constant factors}.

\noindent\textbf{Negative Results:} In the IDExp algorithm, items were chosen for exploration with a probability inversely proportional to their degree. This choice is somewhat non-intuitive -- a more natural choice would seem to be to bias towards higher degree items (as they can reward more users in the universal rewards setting). However, it turns out that a sampling distribution which is proportional to \emph{any other polynomial in the degree} in fact has vanishing competitive-ratio.
\begin{theorem}
\label{thm:negative}
Given any algorithm $\mathcal{A}$ which choses item $i$ for exploration with probability proportional to $d_i^{-1\pm \epsilon}$ for any $\epsilon>0$, then $\exists$ a sequence of graphs $G_n$ (with $Z_{\max}(G_n)=2$), and a corresponding collection of item values $V\in\setR_+^{n_I}$, such that for $r=1$,the competitive-ratio $\gamma^{\mathcal{A}}(G_n,1)$ goes to $0$. 
\end{theorem}

\noindent The formal construction of $G_n$ and proof is presented in Appendix \ref{appsec:conv}. We note that for the graph family $G_n$ used the above result, IDExp achieves a competitive-ratio of $\frac{r}{8e}$.

Finally, in all our algorithms, we split the recommendation slots between explore and exploit recommendations uniformly at random. It is not clear if we need this randomization -- however we can show that some simple intuitive schemes for deciding between explore and exploit are non-competitive.
\begin{theorem}
\label{thm:negexploit}
Suppose we are given a recommendation algorithm $\mathcal{A}$ where the exploit/explore decision based on one of the following rules:
\begin{itemize}[nolistsep,noitemsep]
\item \emph{Exploit-when-possible}: exploit whenever there is a non-$0$ valued available item, else explore. 
\item \emph{Exploit-above-threshold-$t$}: Exploit when the best available item gives a reward $>t$ for some fixed threshold $t>0$.
\end{itemize}
Then, independent of the choice of exploration policy, $\exists$ a sequence of graphs $G_n$ (with $Z_{\max}(G_n)=2$), and corresponding collection of item values $V\in\setR_+^{n_I}$, such that for $r=1$, the competitive-ratio $\gamma^{\mathcal{A}}(G_n,1)$ goes to $0$. 
\end{theorem}

\subsection{Proof Outlines of Converse Results:}
\label{ssec:convoutline}

The main technique we use to obtain converse results is Yao's minimax principle (see \cite{MotwaniRaghavan97}): essentially it states that the competitive-ratio of the \emph{optimal deterministic algorithm} for a given \emph{randomized input} (where the measure over inputs is known to the algorithm) is an upper bound for the competitive-ratio. 

In case of Theorems \ref{thm:conv1}, \ref{thm:convr}, the underlying graph is the complete bipartite graph on $n_U\times n_I$ nodes: now for an $i$ chosen uniformly at random from $N_I$, we set $V(u,i)=1\,\forall\,u\in\mathcal{N}(i)$, and $V(u,i)=0$ for all other $(u,i)$ pairs. Note that the above choice implies that the reward-function $V$ is a \emph{binary uniform} reward-function. Theorem \ref{thm:convr} is more involved -- essentially we show that the competitive-ratio is bounded above by that of an easier `search' problem, where an adversary chooses an item-node, and the users sample $r$ nodes each to try and discover this chosen node. 

Finally, the negative results in Theorem \ref{thm:negative} require constructing a sequence of graphs $G_n$ with associated reward-functions $V_n$ for which $Z_{\max}(G)$ is constant, but the competitive-ratio for non-inverse-degree sampling rules goes to $0$. For the full proofs, refer to Appendix \ref{appsec:conv}.

\section{Discussion and Extensions}
\label{sec:discussion}

\subsection{Inferring Item Values from Multiple Ratings}
\label{ssec:finite}

First, we have assumed that the algorithm knows the value of an item once it has been explored by at least one user. However, as we mentioned in the Section \ref{sec:intro}, a more general condition would be that once an item is viewed by at least $f$ users, its value is known to the algorithm within a multiplicative factor of $(1\pm\delta(f)),$ for some $\delta(f) \in [0, 0.5)$. This generalizes the case considered so far, which corresponds to $f=1$ and $\delta(1)=0$ -- we now show how we can modify our algorithms to handle this more general setting.

We now define an item to be pre-explored if it has been presented to at least $f$ users -- the set of pre-explored items is still denoted $N_I^{expl}$. To provide competitive-ratio guarantees for this setting, we modify the algorithms as follows:
\begin{itemize}[nolistsep,noitemsep]
\item For every user $u$ (or visit $s$ in the infinite-horizon setting), the algorithm chooses $R_1(u)\sim Binomial(r,\frac{f}{f+1})$ slots for exploration, and the rest for exploitation. 
\item The exploration policies are modified in a natural way so as to allow for items to get explored up to $f$ times (see below).
\item For the exploitation, the algorithm still picks the top items from the pre-explored items, based on the noisy estimates of item-value.
\end{itemize}
Now suppose for a user $u$, its top item $i_1^*(u)$, has been explored by at least $f$ users before $u$ arrives. Further, suppose the user decides to exploit at least one item (i.e., $R_2(u)\geq 1$) then either item $i^*_1(u)$ is chosen, or another item $i'\neq i_1^*(u)$, but such that $V(i')\geq(1-2\delta(f))V^*(i^*_k(u))$. The last statement follows since only then $i'$ has higher value than $i^*_1(u)$. Thus the competitive-ratio is reduced at most by a factor $(1-2\delta(f))$.

We now briefly discuss how the exploration step can be done in the infinite-horizon setting for the ULExp algorithm (Algorithm \ref{alg:ulexp2}) -- the arguments for the finite-population setting are similar. Recall that in the algorithm, during each visit $s$, the visiting-user was presented $R_1(s)$ items chosen uniformly from the set of \emph{latest-items} -- those for which it was the first neighbor. We now modify it as follows: each item has a counter, initialized to $0$, which is incremented whenever a neighboring user makes a visit (note: the item may or may not be presented during the visit). Once the counter reaches $f$, the item is declared to be pre-explored if it had been presented to all its $f$ visiting neighbors, else it is discarded. Finally, during each visit $s$, given $R_1(s)$ slots for exploration, the algorithm first chooses $R_1(s)$ numbers $\{l_1,l_2,\ldots,l_{R_1(s)}\}$ from the set $\{0,1,\ldots,f-1\}$ uniformly at random with replacement, and then, for each $l_j$, chooses an item uniformly at random from amongst neighboring items whose counter equals $l_j$. It is easy to see that for $f=1$, this is precisely uniform latest-item exploration. We call this the ULExp-$f$ algorithm, and we have the following theorem:

\begin{theorem}
\label{thm:fexplore}
In the infinite-horizon setting, given graph $G$, users and items arriving according to $Poisson(1)$ processes, and given that for any item, its value is known to within $(1\pm\delta(f))$ after it is explored $f$ times; then the ULExp-$f$ algorithm guarantees:
\begin{equation*}
\gamma^{\mbox{\tiny ULExp-$f$}}(G,r)\geq\frac{1}{(f+1)^{f+1}}\cdot\left(\frac{r}{5Z_{\max}+2}\right)^{f}\cdot(1-2\delta(f))
\end{equation*}
\end{theorem}

Note that substituting $f=1$ and $\delta(f)=0$ gives us back the result from Theorem \ref{thm:bser}. An analogous theorem holds for the finite population case; we skip details for brevity.

\proof{Proof Outline:}
By expanding the latest-item set to items which have seen $<f$ neighboring-user visits, we show that the (expected) size of the latest item set (for the top item corresponding to visit $s$) can now be bounded by $f\cdot(5Z_{\max}+2)$ (where $(5Z_{\max}+2)$ is the bound on the latest-item set for $f=1$ which we derive in the proof of Theorem \ref{thm:ulexp2}). Thus for any visit, its  item is explored by \emph{all of the first $f$ neighboring users} with probability at least $\left(\frac{fr}{f+1}\cdot\frac{1}{f(5Z_{\max}+2)}\right)^f$. Furthermore, the user corresponding to $s$ exploits with probability $\frac{1}{f+1}$, and if her top item $I^*_1(s)$ is pre-explored, it is either presented, or substituted by another pre-explored item with a true value greater than $(1-2\delta(f))V(I^*_1(s))$. Combining these, we get the result. The formal proof is given in Appendix \ref{appsec:multi}.
\Halmos
\endproof

\subsection{More General Reward Models}
\label{ssec:pred}

In the paper till now, we have mostly focused on the \emph{universal rewards} scenario, wherein the reward given by item $i$ to all neighboring user is $V(i)$. This model is studied for ease of exposition and notation; however our proofs allow for more general reward-functions:\\
\noindent\textbf{Personalization:} Item $i$ has intrinsic value $V(i)$, but gives neighboring user $u$ a reward of $V(u,i)=f_{ui}(V(i))$, where $f_{ui}(\cdot)$ are (non-negative, invertible) functions known to the algorithm. This can capture different preferences a user may have vis-\`{a}-vis different items.\\
\noindent\textbf{Collaborative Ranking:} In several setting, the reward earned due to recommending an item may not be possible to quantify -- however, the algorithm can still succeed if it is able to infer a \emph{ranking} of the explored items from user-feedback. This is reminiscent of the Secretary problem (\cite{GenSec}), and also allows for techniques such as in (\cite{jagabathula08}).\\
\noindent\textbf{Probabilistic Predictability:} In many cases, we may be only able to identify the top item for a user \emph{with some probability $P_{pred}$}; for example, in collaborative filtering algorithms such as matrix completion (\cite{keshavan10}). In this case, all our competitive-ratio bounds get scaled by $P_{pred}$.

\section{Related Work}
\label{sec:relwork}

\noindent\textbf{Static Recommendation:} Learning from feedback in large-scale settings is far from being a new idea; however most of the work in this space does not capture user-item dynamics and the explore-exploit tradeoff. Instead, the dominant view is one of taking the user feedback data as a static given (\cite{keshavan10,jagabathula08}); in contrast, our model captures the fact that there is a selection to be made of what user data can be collected, and this selection affects the performance. 

\noindent\textbf{Bandit Algorithms:} Bandit models refer to settings where choosing an action (or \emph{arm}) from a set of actions both yields a reward as well as some feedback about the system, which then affects future control decisions. There are two broad classes of problems that go under the title of bandit problems -- finite-time bandits and infinite-horizon (or Markovian) bandits. 

Markovian bandits (Eg. \cite{Gittins}) focus on settings where each arm has an underlying state, and playing an arm results in a reward that depends on the state, as well as a possible transition to another state (the other arms remaining unaffected). Both rewards and state transition matrices are assumed to be known, and the aim is to maximize the discounted sum reward. Our work differs in that we want to avoid assuming an underlying stochastic model for item-values.

Finite-time bandit problems were originally proposed by \cite{LaiRob} -- subsequent works have greatly generalized the setting by considering different reward-generation processes. Algorithms for these settings control the additive loss (or \emph{regret}) w.r.t. the best policy by bounding the number of times a suboptimal action is chosen. These bounds are in terms of some increasing function of the number of users (plays); however, in content-rich settings where the number of content pieces is of the same order as the number of content-views, it is infeasible that all arms get shown more than a constant number of times. Thus using existing bandit algorithms for our problem leads to a $0$ competitive ratio. For example, consider a setting with $n$ users, $n$ items and a complete bipartite access graph. Suppose one item has a value of $1$ and the rest $0$, and each user is presented with $1$ item (i.e., $r=1$) -- then a bandit algorithm will sample all items at-least once (in particular, the standard UCB algorithm of \cite{Aueretal02a} will sample each arm once just during initialization), thereby getting a competitive-ratio of $\gamma=O(1/n)\rightarrow 0$. On the other hand, the algorithms we present in this work achieve a competitive-ratio of $\frac{1}{8}$.

A notion of an access graph is incorporated in some bandit models such as the Contextual Bandits (\cite{ContBandits}) or Sleeping Bandits (\cite{kleinberg2010regret}) models, the graph and user dynamics are assumed arbitrary. The graph is not used to inform the algorithm design except in that it constrains what items can be shown -- essentially this corresponds to having arbitrary access-constraints, which leads to the results being pessimistic. In our setup, on the other hand, imposing natural stochastic assumptions on user/item dynamics leads to much stronger competitive-ratio guarantees.

\noindent\textbf{Online Matching and its Variants:} Although having the appearance of a bandit problem, our setting is in fact much closer in spirit to certain online optimization problems on graphs. 
Online auction design problems (\cite{Mehtaetal05}) incorporate the fact that an item can be displayed to multiple users, constrained by an underlying graph. However, in such problems the node weights (bids) are known, which often allows greedy algorithms to be constant-factor competitive. Related problems include the generalized secretary problem (\cite{GenSec}) and online transversal-matroid selection (\cite{DimitrovPlaxton}); both are based on a bipartite graph between a `static set' and an `online set of nodes', where node weights (of online nodes) are automatically revealed upon arrival. In contrast, in our problem, the reward-function is unknown, and becomes known only via exploration -- this may affect many future users in a non-trivial manner.

\section{Proofs of Competitive-Ratio Guarantees}
\label{sec:proofs}

We now give the proofs for our results. First, we briefly recall some definitions from before. In the \emph{finite-population model}, we are given graph $G(N_U,N_I,E)$ users $N_U$ and items $N_I$ ($|N_U|=n_U$, $|N_I|=n_I$). For a user $u\in N_U$, we define $\mathcal{N}(u)\triangleq\{i\in N_I|(u,i)\in E\},d_u=|\mathcal{N}(u)|$ (similarly for items). In the infinite-horizon model, the definitions essentially remain the same except that instead of the set of items we have the set of item-classes $N_C$ (where $|N_C|=n_C$). Further, we can define the neighborhood of a user-visit $s$ as items of neighboring classes currently in the system (and similarly for items). 

In the finite-population model, when a user arrives, the recommender algorithm $\mathcal{A}$ presents $r$ items $\{i^{\mathcal{A}}_1(u)\ldots i^{\mathcal{A}}_r(u)\}\subseteq\mathcal{N}(u)$ -- given reward-function $V$, such that the user $u$ earns a reward of $V(u,i)$ from item $i$ (or $V(i)$ if the reward is the same for all users; see Section \ref{sec:discussion}) , the total reward earned by $u$ is $\sum_{k=1}^rV(u,i_k^{\mathcal{A}}(u))$. Further, for a given user $u$, we define the ordering $\{i^*_k(u)\}_{k=1}^{d_u}$ of its neighboring items sorted in decreasing order of their values. Then the algorithm's competitive-ratio (for graph $G$, and for $r$-recommendations per user) is $\gamma^{\mathcal{A}}(G,r)=\inf_{V}\inf_{u\in N_U}\frac{\EE\left[R^{\mathcal{A}}_r(u)\right]}{R_r^*(u)}$. Note that the expectation here is over randomness in user arrival-pattern and the algorithm $\mathcal{A}$; note also that $R^*_r(u)$ \emph{is not random} given $G$ and reward-function $V$. For the infinite-horizon setting, again the definitions are similar, but instead of users, we consider visits -- we will then have $\gamma^{\mathcal{A}}(G,r)=\inf_{V}\inf_{s\in \setN_+}\EE\left[\frac{R^{\mathcal{A}}_r(s)}{R_r^*(s)}\right]$ -- here even the optimal reward is a random variable.

Finally, we use $\setR_+$ for the sets of non-negative reals ($x\geq 0$), and $\setN_+$ for natural numbers ($x\in\{1,2,\ldots\}$). For any $n\in\setN_+$, we define $[n]=\{1,2,\ldots,n\}$. We use the shorthand $a\vee b=\max\{a,b\}$, $a\wedge b=\min\{a,b\}$. {\rev We use $\mathds{1}_{\mbox{E}}$ to denote an indicator random-variable for an event $E$, taking value $1$ when $E$ occurs, else returning $0$ -- similarly we use $\mathds{1}_{\mbox{E}}^{\mathcal{A}}$ to be an indicator r.v. for event $E$ under algorithm $\mathcal{A}$.}

\subsection{A Preliminary Lemma}
\label{ssec:prelim}

We first state and prove a lemma which we use in all our proofs -- it encapsulates the idea that in order to be competitive, it is sufficient to ensure that for every user, \emph{with a near-equal probability, the algorithm should recommend its corresponding highest-valued item}. For ease of exposition, we state the lemma for the finite-population setting. For the infinite-horizon setting, we can get an identical result with user $u$ replaced with user-visit $s$, and \emph{conditioned on the  items currently in the system during visit $s$}.

Given algorithm $\mathcal{A}$ and reward-function $V$, for any pair $(u,i)$ where $i\in\mathcal{N}(u)$ we define $\mathds{1}_{u\rightarrow i}^{\mathcal{A}}$ to be an indicator random variable that is $1$ if user $u$ is shown item $i$ by algorithm $\mathcal{A}$, and else $0$. Then we have:
\begin{lemma}
\label{lem:minind}
Given a graph $G$ and reward-function $V$, then for any algorithm $\mathcal{A}$ displaying $r$ items, we have:
\begin{equation*}
\EE[R_{r}^{\mathcal{A}}(u)]\geq\left(\inf_{u\in N_U}\min_{1\leq k\leq r}\EE\left[\mathds{1}_{u\rightarrow i^*_k(u)}^{\mathcal{A}}\right]\right)R_{r}^*(u)
\end{equation*}
\end{lemma}

\proof{Proof.}
For \emph{non-negative} rewards {\rev (i.e., $V(i)\geq 0\fall i$) }, we can bound the reward earned by user $u$ under algorithm $\mathcal{A}$ as:
\begin{align*}
\EE[R^{\mathcal{A}}_r(u)]&\geq\EE\left[\sum_{k\in[r]}V(i^*_{k}(u))\mathds{1}_{u\rightarrow i^*_k(u)}^{\mathcal{A}}\right]
= \sum_{k\in[r]}V(i^*_{k}(u))\EE\left[\mathds{1}_{u\rightarrow i^*_k(u)}^{\mathcal{A}}\right]\\
&\geq \left(\min_{k\in[r]}\EE\left[\mathds{1}_{u\rightarrow i^*_k(u)}^{\mathcal{A}}\right]\right)\sum_{k\in[r]}V(i^*_{k}(u))=\left(\min_{k\in[r]}\EE\left[\mathds{1}_{u\rightarrow i_k^*(u)}^{\mathcal{A}}\right]\right)R_{r}^*(u)
\end{align*}
Taking infimum over $u\in N_U$ (or $s\in\setN_+$ in the infinite-horizon case), we get the result.
\Halmos
\endproof

\subsection{Performance Analysis: Finite-Population Setting}
\label{ssec:finproof}

Before presenting our proofs, we recall that our algorithms share the following structure:
\begin{itemize}[noitemsep,nolistsep]
\item[$\bullet$] Divide the $r$ recommendations uniformly at random between \emph{explore} and \emph{exploit} recommendations (i.e., the number of exploration slots $R_1\sim Binomial(r,\frac{1}{2})$, rest are for exploitation). 
\item[$\bullet$] For the exploitation step, the algorithm leverages our assumption that for any user, the highest-valued \emph{pre-explored} items can always be identified.
\item[$\bullet$] For the exploration step, we proposed $3$ different exploration policies (in Algorithms \ref{alg:bser},\ref{alg:ider} and \ref{alg:ulexp2}). These are designed to leverage the graph topology and randomness in user-arrivals to ensure \emph{balanced exploration}: for \emph{any neighboring user-item pair}, we can lower-bound the probability that the item is explored before the user arrives to the system.
\end{itemize}
\noindent We also need one additional definition: in the  finite-population setting, we define a \emph{user-arrival pattern} to be a permutation $\pi\in S_{n_U}$ (where $S_{n_U}$ is the set of permutations of users $N_U$) -- we assume that $\pi$ is chosen \emph{uniformly at random}.

\noindent\textbf{Performance Analysis for BPExp Algorithm:} 

\proof{Proof of Theorem \ref{thm:bser}.}
Suppose we are given a reward-function $V$, and a user-arrival pattern $\pi\in S_{n_U}$ chosen uniformly at random. For any user $u$, recall $R_1(u)$ is the number of items explored by $u$; {\rev further, we assume that exploration occurs according to chosen balanced item-partition $M$ (note that $M$ is not random -- \emph{c.f.} Algorithm \ref{alg:bser}). Now let $p_{ui}$ denote the probability that $u$ explores $i$ -- for this to happen, we need that $(u,i)\in M$ (i.e., the edge $(u,i)$ is present in the balanced item-partition which we choose), and further, that $i$ is one of the $R_1(u)$ items explored by $u$.} From the definition of the BPExp algorithm and the makespan $d^*(G)$, conditioned on the events $(u,i)\in M$ and $R_1(u)=k$, {\rev a standard picking-without-replacement argument gives that $u$ explores $i$ with probability at least $\left(\frac{k}{d^*(G)}\wedge 1\right)$.} Thus, we have that \emph{for any neighboring user-item pair $(u,i)$}, $p_{ui}\geq\left[\frac{R_1(u)}{d^*(G)}\wedge 1\right]\cdot\mathds{1}_{\{(u,i)\in M\}}$ -- {\rev note this is a r.v., depending on $R_1(u)$.}

For any item $i\in N_I$, let $u_M(i)$ be the (unique) user connected to it in the item-partition $M$. Recall we define $\mathds{1}_{u\rightarrow i}^{\mbox{\tiny BPExp}}$ to be the indicator that user $u$ is shown (neighboring) item $i$ by algorithm BPExp. Then for any user $u\in N_U$, and any item $i^*_k(u), k\in[r]$ (i.e., one of the top $r$ items for $u$), we have {\rev that $\mathds{1}_{u\rightarrow i^*_k(u)}^{\mbox{\tiny BPExp}}=1$ iff}:
\begin{enumerate}[noitemsep,nolistsep]
\item[i.] $u=u_M(i^*_k(u))$ AND $u$ explores $i^*_k(u)$, 
\\OR
\item[ii.] $u_M(i^*_k(u))=u'\neq u$ AND $u'$ arrives before $u$ in arrival-pattern $\pi$ AND $u'$ explores $i^*_k(u)$.
\end{enumerate}
The first condition captures the case where $u$ \emph{explores} $i^*_k(u)$ (via the $R_1(u)$ slots used for exploration). The second condition captures the case where $i^*_1(u)$ is explored by the time $u$ arrives, and hence $u$ can \emph{exploit} it via its $R_2(u)$ exploration slots. Note that the above options are \emph{mutually exclusive}; which of the conditions holds is uniquely determined given values $V$ and the item-partition $M$. Now under condition $i$, {\rev using our previous characterization of $p_{ui}$ and Jensen's inequality}, we have that: 
\begin{align*}
\EE\left[\mathds{1}_{u\rightarrow i^*_k(u)}^{\mbox{\tiny BPExp}}\right] \geq\EE\left[\EE\left[\left.\frac{k}{d^*(G)}\wedge 1\right|R_1(u)=k\right]\right]\geq\frac{r}{2d^*(G)}\wedge 1
\end{align*}
\noindent Under condition $ii$, by a similar calculation, we have that the probability of $u'$ exploring $i^*_k(u)$ is $\geq\left(\frac{r}{2d^*(G)}\wedge 1\right)$. Further, \emph{since $\pi$ is chosen uniformly at random from $S_{n_U}$}, we have that $u'$ arrives before $u$ with probability $1/2$ {\rev(more generally, for any two users $u,v\in N_U$, $u$ arrives before $v$ with probability $1/2$). Note that the expected reward under the second condition is lower than that in the first case -- since the two are mutually exclusive, a lower bound on the performance under the second condition translates to a lower bound for the BPExp algorithm.}

{\rev Finally, to bound the performance of BPExp under condition $ii$, we observe that it can be stochastically under-dominated via the following modified algorithm: First, note that choosing $R_1\sim Binomial(r,\frac{1}{2})$ is equivalent to sequentially allocating slots $\{1,2,\ldots,r\}$ to either exploration with probability $1/2$, else exploitation. Now when user $u$ arrives, suppose we allocate $R_2(u)$ slots $\{k_1,k_2,\ldots,k_{R_2}\}\subseteq [r]$ for exploitation -- then instead of showing the top $R_2(u)$ pre-explored items, we show items $\{i^*_{k_1}(u),i^*_{k_2}(u),\ldots,i^*_{k_{R_2}}(u)\}$ if they have been pre-explored, and fill any remaining exploitation slot with the top remaining pre-explored items. A coupling argument shows that this modified algorithm is stochastically dominated by BPExp (since it may recommend a less valuable item). However, under the modified policy, it is easy to see that $\fall k\in[r]$, whenever item $i^*_k(u)$ is in the set of pre-explored items, then user $u$ exploits it with probability $1/2$ -- thus, under condition $ii$ using the modified policy, we have:
\begin{align*}
\EE\left[\mathds{1}_{u\rightarrow i^*_k(u)}\right] &=\EE\left[\mathds{1}_{\{u'\mbox{ arrives before }u\}}
\mathds{1}_{\{u'\mbox{ explores }i^*_k(u)\}}
\mathds{1}_{\{u\mbox{ exploits }i^*_k(u)\}}\right]\\
&\geq\frac{1}{2}\cdot\frac{1}{2}\cdot\left(\frac{r}{2d^*(G)}\wedge 1\right)=\left(\frac{r}{8d^*(G)}\wedge \frac{1}{4}\right)
\end{align*}

Finally, using Lemma \ref{lem:minind}, and taking infimum over users $u\in N_U$, we get the result,
}
\Halmos
\endproof

\noindent\textbf{Performance Analysis for IDExp Algorithm:} Next, we prove Theorem \ref{thm:ider}; refer Section \ref{ssec:ide} for details and theorem statement. We first need a lemma characterizing Greedy Neighborhood Partitioning (Definition \ref{def:greedypartn}):
\begin{lemma}
\label{lem:balance}
If $\{P_u\}_{u\in N_U}$ is generated by independently applying Greedy Neighborhood Partitioning for each user $u\in N_U$, then $Z_{max}(G,r,\{P\})\triangleq\max_{u\in N_U}\max_{k\in[r]} \sum_{i\in P_u^k}d_i^{-1}$ obeys:
\begin{equation*}
Z_{max}(G,r,\{P\})\leq \frac{2Z_{max}(G)}{r}.
\end{equation*}
\end{lemma}

\proof{Proof.} 
First, note that by definition, $\sum_{i\in\mathcal{N}(u)}d_i^{-1}$ is bounded by $Z_{\max}(G)$ for all $u$. Further, $d_i^{-1}\leq 1$ for all $i$. Together, this implies that the \emph{optimal} balanced neighborhood partition $\{\tilde{P}_u^k\}_{k\in[r]}$ for any user $u$ has the property that $\max_{k\in[r]} \sum_{i\in \tilde{P}_u^k}d_i^{-1}\leq \frac{Z_{\max}(G)}{r}$. The above claim now follows from the existing result of \cite{Graham66}, which shows that greedy set partitioning has an approximation ratio of $\frac{1}{2}$.
\Halmos
\endproof

Using this bound, we can now prove the stated result.

\proof{Proof of Theorem \ref{thm:ider}.}
Consider any user $u$. First, from Lemma \ref{lem:minind}, we get:
\begin{equation*}
\EE[R_r^{\mbox{\tiny IDExp}}(u)]\geq\left(\min_{k\in [r]}\EE\left[\mathds{1}_{u\rightarrow i_k^*(u)}^{\mbox{\tiny IDExp}}\right]\right)\cdot R_{r}^*(u)
\end{equation*}
As before, we drop the superscript indicating that relevant quantities are conditional on using the IDExp algorithm. We now show that the algorithm results in a uniform lower bound \emph{over all users} $u\in N_U$, and $\fall k\in[r]$, of $\EE\left[\mathds{1}_{u\rightarrow i^*_k(u)}\right]\geq\frac{1}{4e}\left(\frac{r}{2Z_{\max}(G)}\wedge 1\right)$; substituting this in the above equation, we get our result.

{\rev The difficulty in analyzing IDExp is that the item-explorations are no longer independent, but rather, depend on the decisions made by all previous users. However, we can stochastically dominate the IDExp algorithm by a fictitious algorithm wherein for each user, \emph{all its neighboring items are eligible for exploration}, irrespective of whether they have been explored before. Clearly this can only make the performance worse, as under our assumption that an item's value is known once explored.} Further, we define $Z=\frac{2Z_{\max}(G)}{r}\vee 1$. Now, for any user $u$ and neighboring item $i$, we claim that the probability that $u$ explores $i$, given by $p_{ui}$, obeys:
\begin{equation*}
p_{ui}\geq \frac{d_i^{-1}}{2Z}
\end{equation*} 
To see this, first note that from Lemma \ref{lem:balance}, we have that for every user $u$, and every neighborhood partition $P_u^k, k\in[r]$, we have that $\sum_{i\in P_u^k}d_i^{-1}$ is bounded by $Z$. Now for user $u$, the number of explore slots is $R_1(u)\sim Binomial(r,1/2)$ -- thus $p_{ui}\geq\EE\left[\frac{R_1(u)}{rZ}\right]=\frac{d_i^{-1}}{2Z}$. 

Now given reward-function $V$ and user arrival-pattern $\pi$ chosen uniformly at random from $S_{n_U}$, consider any user $u\in N_U$ with associated highest-valued $r$ items $i^*_k(u), k\in[r]$. Consider item $i_k^*(u)$ -- let $A_t$ be the event that there are $t\in\{0,1,\ldots,d_i-1\}$ neighbors of $i_k^*(u)$ who arrive in the system before $u$ in arrival-pattern $\pi$; we denote these users as $\{a_k\}_{k=1}^t$. Conditioned on $A_t$, under the IDExp algorithm, $\mathds{1}_{u\rightarrow i_k^*(u)}=1$ iff $i_k^*(u)$ is explored by $u$ OR by explored by one of the $t$ neighbors of $i_k^*(u)$ who arrived before $u$, and exploited by $u$. As in the proof of Theorem \ref{thm:bser}, we have that the probability that $i_k^*(u)$ is exploited by $u$ when it is pre-explored is at least $\frac{1}{2}$. Thus we have (using $i$ as shorthand for $i_k^*(u)$):
\begin{align*}
\EE\left[\left.\mathds{1}_{u\rightarrow i_k^*(u)}\right|A_t\right]
&\geq\frac{1}{2}\left[1-(1-p_{ui})\Pi_{k=1}^t(1-p_{a_ki})\right]\\
&\geq\frac{1}{2}\left[p_{a_1i}+(1-p_{a_1i})p_{a_2i}+(1-p_{a_1i})(1-p_{a_2i})\ldots(1-p_{a_ti})p_{ui}\right].
\end{align*}
Note that $\frac{d_i^{-1}}{2Z}\leq p_{ui}\leq\frac{1}{d_i}$. Now we have:
\begin{align*}
\EE\left[\left.\mathds{1}_{u\rightarrow i_k^*(u)}\right|A_t\right]
&\geq\frac{1}{2}\sum_{k=0}^{t}\left(1-\frac{1}{d_i}\right)^k\frac{d_i^{-1}}{2Z}
=\frac{1}{2Z}\left(1-\left(1-\frac{1}{d_i}\right)^{t+1}\right)
\end{align*}
Since $\pi$ is drawn uniformly at random from $S_{n_U}$, we have that $\PP[A_t]=\frac{1}{d_i}$. Thus:
\begin{align*}
\EE[\mathds{1}_{u\rightarrow i_1(u)}]&\geq\frac{1}{d_i}\cdot \frac{1}{4Z}\sum_{t=1}^{d_i}\left(1-\left(1-\frac{1}{d_i}\right)^{t}\right)\\
&= \frac{1}{4Z}\left(\frac{1}{d_i}+\left(1-\frac{1}{d_i}\right)^{d_i+1}\right)\\
&\geq \frac{1}{4Z}\left(\frac{1}{d_i}+\frac{1}{e}\left(1-\frac{1}{d_i}\right)^2\right),
\end{align*}
since $\left(1-\frac{1}{d}\right)^d\geq\frac{1}{e}-\frac{1}{ed}\fall d\in \setN_+$. Thus we have:
\begin{align*}
\EE[\mathds{1}_{u\rightarrow i_1^*(u)}]
&\geq \frac{1}{4Z}\left(\frac{1}{e}+\frac{1}{d_i}\left(1-\frac{2}{e}\right)+\frac{1}{ed_i^2}\right)\geq\frac{1}{4eZ},
\end{align*}
where $Z=\frac{2Z_{\max}(G)}{r}\vee 1$. This completes the proof.
\Halmos
\endproof

\subsection{Performance Analysis: Infinite-Horizon Setting}
\label{ssec:infproof}

Finally, we turn to the infinite-horizon setting. Recall that we now have a graph between users $N_U$ and item-classes $N_C$, with user-visits $x\in\setN_+$ and items $i\in\setN_+$. Each item $i$ has an item-class $C(i)$ (according to the underlying independent Poisson processes), a lifetime $\tau$ (same for all items) and a reward-function $V(i)$. More specifically, for each item-class $c$, we define $V_c$ to be an arbitrary, infinite sequence of reward-functions, such that the $k^{th}$ item of class $c$ has the $k^{th}$ reward function in the sequence (i.e., $V_c(k)$).

We define $S_1(i)$ to be the first visit by a neighboring user $u\in\mathcal{N}(C(i))$ after item $i$ arrives to the system (and before it expires). Complementary to this, for visit $s$, we defined the latest-items set $L(s)$. Finally the Uniform Latest-Item Exploration strategy is based on randomly picking $R_1(s)\sim Binomial(r,1/2)$ items from $L(s)$ without replacement, for exploration during visit $s$.

The main idea behind latest-item exploration is that it can be shown that for any \emph{typical} item, the expected size of the latest-items set is bounded by $2Z_{\max}(G)+1$. Coupled with Jensen's inequality, this result suggests that the probability that any typical item is explored is greater than $\frac{1}{2Z_{\max}(G)+1}$ -- however this is not sufficient to obtain our result because for a given user, we are interested in the latest-item set \emph{as seen by its corresponding highest-valued items}. This however is determined by the reward-function $V$, and it is not clear how the dependence can be quantified. This is the main technical challenge in the proof.

The main idea of our proof is as follows -- for any visit $s$ arriving at time $T_s$, we consider the arrivals in the time-interval $[T_s-\tau,T_s]$ -- since each item has a lifetime of $\tau$, the items in this interval are the only ones which matter. We argue that the statistics of these arrivals are unaffected by the reward-function $V$, and further, using the ULExp scheme, we can control the probability with which a particular item is explored. For this, we first require the following combinatorial lemma:

\begin{lemma}
\label{lem:redblue}
Given a uniform permutation of $R$ red and $B$ blue balls, let $N_{cons}(i),\, i\in[R]$ be the number of consecutive red balls (i.e., bounded on either side by either a blue ball or a boundary) containing the $i^{th}$ red ball. Then we have: 
\begin{equation*}
\max_i\EE[N_{cons}(i)]\leq \frac{4R}{B+1}+2
\end{equation*} 
\end{lemma}
\noindent\textbf{Remarks:} For ease of notation, we define $N(R,B)=\max_i\EE[N_{cons}(i)|R\mbox{ red},B\mbox{ blue balls}]$. $N(R,B)$ is clearly greater than the expected number of consecutive balls (which, by symmetry is $R/(B+1)$); a more subtle fact is that $N(R,B)$ is greater than the expected number of consecutive red balls \emph{as seen by a random red ball} (unlike, for example, the PASTA property for queueing processes). This is due to the presence of boundary conditions -- for example, for $B=1$, we can compute the expected consecutive sequence seen by a random ball to be $(2R+1)/3$, while we show in the proof below that $N(R,B)$ in this case is $(3R+1)/4$. Crucially however, our bound on $N(R,B)$ is much less than the expected value of the \emph{maximum number of consecutive balls} -- in the case where $B=R=n$, it is known that the longest sequence is $\Theta\left(\log n/\log\log n\right)$, while our bound on $N(n,n)$ is $6$.

\proof{Proof.}
First, note that when $B\leq 3$, the bound given in the lemma evaluates to a value $\geq R$, which is a trivial upper bound on the length of a consecutive subsequence. Hence, we essentially need to prove it for $B>3$ and general $R$ -- further, in this range, we can lower bound the RHS by $\frac{4(R+1)}{B+1}+1$, which is the bound we will prove. The proof outline is as follows -- first we exactly evaluate the quantity $N(R,1)$ (i.e., for the case $B=1$) -- subsequently we use an induction argument, wherein we bound $N(R,B)$ by a function of $N(R,B-1)$. We show that this function is increasing as long as $B>3$, and then verify the inequality inductively.

First we explicitly compute $N(R,1)$. Here, it is easy to see that the index of the red ball which sees the longest expected consecutive sequence is $i^*=\lceil R/2 \rceil$ -- for any other index $i$, the number of consecutive red balls is either the same (if the lone blue ball falls on the same side of $i$ and $i^*$) or less (if it falls in between). Now in case $R$ is odd, we have:
\begin{align*}
N(R,1)=\frac{1}{R+1}\left(2.\frac{R+1}{2}+2.\frac{R+3}{2}+\ldots+2.R\right)=\frac{3R+1}{4}
\end{align*}
Similarly, in case $R$ is even, we have:
\begin{align*}
N(R,1)&=\frac{1}{R+1}\left(\frac{R}{2}+2.\frac{R+2}{2}+\ldots+2.R\right)\leq\frac{3R+1}{4}
\end{align*}
Thus, combining the two cases, we get $N(R,1)\leq \frac{3R+1}{4}$. 

Now to obtain the bound on $N(R,B)$ for larger values of $B$, we bootstrap the above result as follows: Suppose we define $X(R,B-1,i)$ to be the length of the consecutive sequence as seen by the $i^{th}$ red ball after we drop $B-1$ blue balls uniformly at random. Then from the above argument, we have:
\begin{align*}
\EE[X(R,B,i)|X(R,B-1,i)=k]&\leq\frac{k+1}{R+1}.\frac{3k+1}{4}+\left(1-\frac{k+1}{R+1}\right).k\\
&=\frac{-k^2+4(R+1)k+1}{4(R+1)},
\end{align*}
and taking expectations, via Jensen's inequality, we get:
\begin{align*}
\EE[X(R,B,i)]&\leq\frac{-\EE[X(R,B-1,i)]^2+4(R+1)\EE[X(R,B-1,i)]+1}{4(R+1)},
\end{align*}
Now note that $f(k)=-k^2+4(R+1)k+1$ is increasing for $k\leq 2 (R+1)$ -- hence we can further upper bound the RHS by replacing $\EE[X(R,B-1,i)]$ by $N(R,B-1)$. Finally, since the resulting expression is independent of $i$, we can replace $\EE[X(R,B,i)]$ with $N(R,B)$. Rearranging the above  inequality, we have:
\begin{align*}
N(R,B)&\leq N(R,B-1)+\frac{1-N(R,B-1)^2}{4(R+1)},
\end{align*}
Finally, suppose we have that $N(R,B-1)$ satisfies the required bound. Again, since the RHS is increasing as long as $N(R,B-1)\leq 2R+2$, thus for $B>3$, we have:
\begin{align*}
N(R,B)&\leq \frac{4(R+1)}{B}+1-\frac{16(R+1)^2/B^2+8(R+1)/B}{4(R+1)}\\
&\leq \frac{4(R+1)}{B}+1-\frac{4(R+1)/B+2}{B}\\
&\leq 1+4(R+1)\frac{B-1}{B^2}\leq 1+\frac{4(R+1)}{B+1}
\end{align*}
This completes the proof.
\Halmos
\endproof

\proof{Proof of Theorem \ref{thm:ulexp2}.}
From Lemma \ref{lem:minind}, we have that for any visit $s$:
\begin{equation*}
\EE\left[R^{\mbox{\tiny ULExp}}_r(s)\right]\geq\EE\left[\min_{k\in[r]}\EE\left[\mathds{1}^{\mbox{\tiny ULExp}}_{s\rightarrow I^*_k(s)}\right]R^*_r(s)\right]
\end{equation*}
\noindent  Here the inner expectation is over the randomness in the algorithm, and the outer expectation is over randomness in the sample path. We henceforth suppress the superscript.

To complete the proof, for any user-visit $s$, and for all $k\in[r]$, we need to show that the corresponding $k^{th}$-top item, denoted $I_k^*(s)$ satisfies $\EE\left[\mathds{1}^{\mbox{\tiny ULExp}}_{s\rightarrow i^*_k(s)}\right]\geq \frac{r}{4(5Z_{\max}(G)+2)}$. We will in fact prove a more general result -- suppose items of any item-class $c$ arrive at rate $\lambda_c$, and each user $u$ visits at rate $\lambda_u$. We now obtain that $\EE\left[\mathds{1}_{s\rightarrow I^*_k(s)}\right]\geq\frac{r}{4(5Z+2)}$, where:
\begin{equation*}
Z=\max_{u\in N_U}\sum_{c\in\mathcal{N}(u)}\frac{\lambda_c}{\sum_{u'\in\mathcal{N}(c)}\lambda_{u'}}
\end{equation*}
When $\lambda_u=\lambda_c=1\fall u,c$, we get the claimed result.

Now suppose we denote $L(I_k^*(s))$ to be the size of the latest-item set of the first-neighbor of item $I_k^*(s)$ (formally, in our notation, $L(S_1(I_k^*(s)))$ -- we use $L(I_k^*(s))$ as a shorthand for this). Then we have that the probability of $I_k^*(s)$ being explored by $S_1(I_k^*(s))$ under the ULExp policy is given by:
\begin{equation*}
\PP[I^*_k(s)\mbox{ is explored by }S_1(I_k^*(s))]=\EE\left[\frac{R_1(S_1(I_k^*(s)))}{|L(I^*(s))|}\right]\geq \frac{r}{2\EE\left[|L(I^*(s))|\right]},
\end{equation*}
where for the last inequality, we have used that $R_1(S_1(I_k^*(s)))$ is independent of $L(I^*_k(s))$, and further bounded it via Jensen's inequality. Further, via similar arguments as in Theorems \ref{thm:bser} and \ref{thm:ider}, we have that: 
\begin{equation}
\label{eq:pvsL}
\EE\left[\mathds{1}_{s\rightarrow i^*_k(s)}\right]\geq \frac{1}{2}  \PP[I^*_k(s)]\mbox{ is explored by  }S_1(I_k^*(s))]\geq\frac{r}{4}\frac{1}{\EE\left[|L(I^*_k(s))|\right]}.
\end{equation}

Thus a lower bound on the competitive-ratio essentially involves upper bounding $\EE\left[|L(I^*_k(s))|\right]=\EE\left[|L(s')||i=I^*_k(s),s'=S_1(i)\leq s\right]$. Note that the conditioning depends on the bipartite-graph $G$, and also, on the reward-function sequence $V$; it can not be removed in a trivial manner (i.e., we can not argue that $\EE\left[|L(I^*_k(s))|\right]=\EE\left[|L(s')|\right]$ for some `typical' visit $s'$). Instead, we need to exactly characterize and bound the dependence on the graph and reward-function. 

We do so as follows: given user $s$ arrives at time $T_s$, we consider all sample-paths of the process parametrized by two sets of random variables:
\begin{itemize}[nolistsep,noitemsep]
\item $\textbf{I}_{l}(s)=\{I_{l}(s,c)\}_{c\in N_C}$ are the indices of the most recent items for each item-class.
\item $\textbf{R}_s=\{R_c\}_{c\in N_C}$ are the number of items of each class that arrived in the interval $[T_s-\tau,T_s]$, and similarly $\textbf{B}_s=\{B_u\}_{u\in N_U}$ are the number of visits by each user in the same time interval.
\end{itemize}
Since all items have a lifetime of $\tau$, it is clear that $I^*_k(s)$ must have arrived in the interval $[T_s-\tau,T_s]$. Further, given $\textbf{I}_{l}(s), \textbf{R}_s$ and $\textbf{B}_s$, $I^*_k(s)$ is deterministic -- we can now define $c^*=C(I^*_k(s))=c^*\in\mathcal{N}(U(s))$, and further $i^*$ to be the index (or position) of $I^*_k(s)$ among all the items of class $c^*$ arriving in the interval (i.e., $i^*\in\{1,2,\ldots,R_{c^*}\}$). The crucial observation is that conditioning on $\{\textbf{R}_s,\textbf{B}_s\}$ item/user-visits
arriving in the interval implies that \emph{any ordering of these $\{\textbf{R}_s,\textbf{B}_s\}$ events is equally likely} -- further, this remains unchanged given $\textbf{I}_{l}(s)$. This now puts us in a position where we can use Lemma \ref{lem:redblue}.

Recall that we want an upper bound on $\EE[|L(I^*_k(s))|]$ -- as we argued, given the conditioning presented above, this corresponds to the item of class $c^*$ with index $i^*$ among all items of that class in the interval $[T_s-\tau,T_s]$. Now we define $L(I^*_k(s),c)$ to be the number of latest items of item-class $c$ encountered by $S_1(I^*_k(s))$ -- thus $L(I^*_k(s))=\sum_{c\in\mathcal{N}(u(s))}L(I^*_k(s),c)$. Note that the first-visit of $I^*_k(s)$ could correspond to any neighboring user; from Lemma \ref{lem:redblue}, we thus have:
\begin{align*}
\EE\left[|L(I^*_k(s),c^*)|\right]&\leq\EE\left[\frac{4R_{c^*}}{1+\sum_{u\in\mathcal{N}(c^*)} B_{u}}+2\right] + \EE[L_{prior}],
\end{align*}
where $L_{prior}$ is the number of additional items which arrived before $T_s-\tau$, but were potentially in $L(I^*_k(s))$. Now note that $R_{c^*}\sim Poisson(\lambda_{c*}\tau)$, and further, for its neighboring users, $\sum_{u\in\mathcal{N}(c^*)} B_{u}\sim Poisson(\sum_{u\in\mathcal{N}(c^*)}\lambda_u\tau)$ (since they are a sum of independent Poisson processes). Thus we have:
\begin{align*}
\EE\left[|L(I^*_k(s),c^*)|\right]&\leq\EE\left[4R_{c^*}\right]\EE\left[\frac{1}{\sum_{u\in\mathcal{N}(c^*)} B_{u}+1}\right]+2+\frac{\lambda_{c^*}}{\sum_{u\in\mathcal{N}(c^*)}\lambda_u}\\
&\leq \frac{4\lambda_{c^*}\tau\left(1-\exp(-\tau\sum_{u\in\mathcal{N}(c^*)}\lambda_u)\right)}{\sum_{u\in\mathcal{N}(c^*)}\lambda_u\tau}+2+\frac{\lambda_{c^*}}{\sum_{u\in\mathcal{N}(c^*)}\lambda_u}\\
&\leq \frac{5\lambda_{c^*}}{\sum_{u\in\mathcal{N}(c^*)}\lambda_u} +2
\end{align*}
\noindent To complete the proof, we need to get a bound on $\EE\left[|L(I^*_k(s),c)|\right]\fall c\neq c^*$. Consider any visit $s'$ in the interval $[T_s-\tau,T_s]$ -- conditioning only on $\{\textbf{R}_s,\textbf{B}_s\}$, it follows from symmetry that 
\begin{equation*}
\EE\left[|L(s')||\{\textbf{R}_s,\textbf{B}_s\}\right]\leq \sum_{c\in\mathcal{N}(U(s'))}\frac{R_c}{1+\sum_{u\in\mathcal{N}(c)}B_u} +\frac{\lambda_c}{\sum_{u\in\mathcal{N}(c)}\lambda_u},
\end{equation*}
where the second term accounts for the arrivals prior to $T_s-\tau$. In our case, we are interested in $L(S_1(I_k^*(s)),c)$ -- so we need to take into account the condition that visit $s'$ saw $I_k^*(s)$ in its latest-item set. However, in case of items of class $c^*$, we have that the number of items in the latest-item set of $S_1(I_k^*(s))$ can at most increase by a factor of $4$. Thus, via similar arguments as above, we can show that:
\begin{equation*}
\EE\left[|L(I^*_k(s),c)|\right]\leq\EE\left[\frac{4R_c}{1+\sum_{u\in\mathcal{N}(c)}B_u}\right]+\frac{\lambda_c}{\sum_{u\in\mathcal{N}(c)}\lambda_u},
\end{equation*}
\noindent and thus we have, for all user-visits $s$ and for all $k\in[r]$:
\begin{align*}
\EE\left[|L(I^*_k(s))|\right]&\leq\EE\left[|L(I^*_k(s),c^*)|\right]+\EE\left[\sum_{c\in\mathcal{N}(S_1(I^*_k(s))),c\neq c^*}|L(I^*_k(s),c^*)|\right]\\
&\leq 2+\EE\left[\sum_{c\in\mathcal{N}(S_1(I^*_k(s)))}\frac{4R_c}{1+\sum_{u\in\mathcal{N}(c)}B_u}+\frac{\lambda_c}{\sum_{u\in\mathcal{N}(c)}\lambda_u}\right]\\
&\leq 2+\max_{u\in N_U}\sum_{c\in\mathcal{N}(u)}\left[\frac{5\lambda_c}{\sum_{u'\in\mathcal{N}(c)}\lambda_{u'}}\right]=2+5Z
\end{align*}
Now can substitute this in Equation \ref{eq:pvsL}, to get the result.
\Halmos
\endproof

\bibliographystyle{ormsv080}
\bibliography{GraphReco}

\begin{APPENDICES}

\section{Converse Results}
\label{appsec:conv}

The main technique we use to show upper bounds on $\gamma$ over \emph{any online algorithms} is Yao's minimax principle (refer \cite{MotwaniRaghavan97}): the competitive ratio of the \emph{optimal deterministic algorithm} for a given \emph{randomized input} is an upper bound for the competitive ratio. Note that the algorithms are aware of the input distribution.

\subsection{Upper Bound: Competitive-ratio for the Complete Bipartite Graph:}

{\rev
\proof{Proof of Theorem \ref{thm:conv1}.}
Consider the complete bipartite graph $\overline{G}(N_U,N_I,E)$, i.e., $E=\{(u,i)\fall u\in N_U,i\in N_I\}$, with $|N_U|=n_U$ and $|N_I|=n_I$. We choose a \emph{single item} $i^*$ uniformly at random from $N_I$ and set $V(i^*)=1$; the remaining items have $V(i)=0$ -- thus $R^*(u)=1$ for any user $u$. We denote the competitive ratio of the best \emph{deterministic} algorithm in this setting to be $\gamma_{det}(\overline{G},1)$. Then by Yao's minimax principle, we have that for any randomized online algorithm that makes a single recommendation:
$\gamma(G,1)\leq\gamma_{det}(\overline{G},1)$.

For a user $u$, let the expected reward achieved by the best deterministic algorithm be denoted $R_{det}(u)$; further, let the expected sum of rewards over all users be $R_{det}$. From symmetry, we it is clear that $R_{det}(u)=\frac{R_{det}}{n_U}\fall u$. We now claim that \emph{the optimal deterministic algorithm achieves}:
\begin{align}
\label{eq:claim}
R_{det}
&=  \frac{\min\{n_U,n_I\}\left(2n_U+1-\min\{n_U,n_I\}\right)}{2n_I}
\end{align}
\noindent Also, note that for any user $u$, $R^*(u)=1$, and thus $\gamma_{det}(\overline{G},1)=\frac{(2n_U+1-\min\{n_U,n_I\})}{(2n_I)\max\{1,n_U/n_I\}}$. On the other hand, we have $Z_{\max}(\overline{G})=\frac{n_U}{n_I}$ -- thus, given $\epsilon>0$, for large enough $n_I$ we have 
\begin{align*}
\gamma(G,1)\leq\frac{1}{2\cdot\max\{1,Z_{\max}(\overline{G})\}}+\epsilon.
\end{align*}

To complete the proof, we establish equation \ref{eq:claim} via a $2$-dimensional induction argument on $(n_U,n_I)$. We denote the LHS of equation \ref{eq:claim} as $R_{det}(n_U,n_I)$. The base cases are easy to establish -- for $(n_U,1)$, equation \ref{eq:claim} gives $R_{det}(n_U,1)= n_U$, and for $(1,n_I)$ we have $R_{det}(1,n_I)=1/n_I$; both these hold trivially for any deterministic algorithm. Now suppose equation \ref{eq:claim} holds for all $(n_U',n_I')$ such that either $n_U'< n_U, n_I'\leq n_I$ or $n_U'\leq n_U, n_I'< n_I$. Now to compute $R_{det}(n_U,n_I)$, we observe that any deterministic algorithm either uncovers item $i^*$ in the first exploration, else it reduces to a complete bipartite graph with $(n_U-1,n_I-1)$ nodes. Since $i^*$ is picked uniformly at random by the adversary, we have:
\begin{align*}
R_{det}(n_U,n_I)
&=\frac{1}{n_I}\cdot n_U + \frac{n_I-1}{n_I}R_{det}(n_U-1,n_I-1),
\end{align*}
\noindent and using the induction hypothesis, we get: 
\begin{align*}
R_{det}(n_U,n_I)&=\frac{1}{n_I}\cdot n_U + \left(\frac{n_I-1}{n_I}\right)\left(\frac{\min\{n_U-1,n_I-1\}\left(2(n_U-1)+1-\min\{n_U-1,n_I-1\}\right)}{2(n_I-1)}\right)\\
&=\frac{2n_U+\left(\min\{n_U,n_I\}-1\right)\left(2n_U-\min\{n_U,n_I\}\right)}{2n_I}\\
&=\frac{\min\{n_U,n_I\}\left(2n_U-\min\{n_U,n_I\}+1\right)}{2n_I}
\end{align*}
This completes the proof.
\Halmos
\endproof
}

\subsection{Upper Bound: Scaling with Number of Recommendations:}

\proof{Proof of Corollary \ref{thm:convr}.}
Consider the complete bipartite graph $\overline{G}(N_U,N_I,E)$ with binary rewards, as in the proof of Theorem \ref{thm:conv1}. $V^*=\{i\in N_I|V(i)=1\}$ is now chosen to be a uniformly-random set of $r$ items from $N_I$; thus the sum of optimal offline rewards $R^*_r(\overline{G})=rn_U$. Let $R_{det,r}(\overline{G})$ be the sum reward earned by the optimal deterministic algorithm showing $r$ items on graph $G$ -- by symmetry, we have that the per-user competitive ratio is the same as the ratio of the total reward earned by the algorithm $R_{det,r}(\overline{G})$ to the total optimal offline reward $R^*$.

Now consider an alternate problem where we have a complete bipartite graph $\overline{G}'(N_U',N_I',E')$, where $N_I'=N_I$ and ${V'}^{*}=V^*$, but $|N_U|=rn_U$ -- essentially, $\overline{G}'$ is derived from graph $\overline{G}$ by making $r$ copies of each user. We henceforth use $\overline{G}$ and $\overline{G}'$ as shorthand to refer to these two problems.

Suppose now in problem $\overline{G}'$, we can recommend only a single item. Then clearly $R^{*}_1(\overline{G})'$ (the optimal offline reward in $\overline{G}'$) is $rn_U$. Let $R_{det,1}(\overline{G}')$ denote the expected reward earned by the optimal deterministic algorithm showing $1$ recommendation on graph $\overline{G}'$. Then we have that $R_{det,1}(\overline{G}')\geq R_{det,r}(\overline{G})$; this follows from the fact that any deterministic algorithm that recommends $r$ items on graph $\overline{G}$ can be converted to a deterministic algorithm for graph $\overline{G}'$ (by recommending to the first $r$ users of $\overline{G}'$ the same $r$ items as recommended to the first user in $\overline{G}$, and so on for each block of $r$ users in $\overline{G}'$) such that they have the same rewards. However the class of all deterministic algorithms for $\overline{G}'$ is larger (in particular, it includes algorithms that recommend the same item to multiple users in a block of $r$ consecutive users, which in the aforementioned mapping would correspond to recommending the same item multiple times to the same user in $\overline{G}$). Now using Theorem \ref{thm:conv1}, we have that:
\begin{align*}
\gamma_{det}(\overline{G},r)\leq\gamma_{det}(\overline{G}',1) &=\begin{dcases}
   \frac{rn_U}{2n_I}+\frac{1}{2n_I} & :rn_U\leq n_I \\
   \frac{1}{2}+\frac{rn_U-n_I+1}{2rn_U}   & :rn_U> n_I 
  \end{dcases},
\end{align*}   
Hence by Yao's minimax principle, for given $\epsilon\geq 0$, $r$ and $n_U$ and for any randomized online algorithm that recommends $\leq r$ items, for sufficiently large $n_I$, we have $\gamma_r\leq\frac{rn_U}{n_I}+\epsilon$. \Halmos
\endproof

\subsection{Negative Result: Non Inverse-Degree Dynamic Exploration}

\proof{Proof of Theorem \ref{thm:negative}.}
We first define a family of graphs: for $n\in\mathbb{Z}^+$, we define $\widehat{G}_n(N_U,N_I,E)$:
\begin{itemize}[noitemsep,nolistsep]
\item[$\bullet$] $|N_U|=n$ and $|N_I|=2n$; each user has an index in $[n]$, and similarly each item an index in $[2n]$. 
\item[$\bullet$] Each user is connected to the item with the same index, i.e., $(j,j)\in E\fall j\in [n]$. 
\item[$\bullet$] The remaining items are connected to all users, i.e., $(u,i)\in E\fall u\in [n],i\in \{n+1,n+1,\ldots,2n\}$
\end{itemize}
One can show that $Z_{\max}(\widehat{G}_n)=2\fall n$; thus recommendation via the IDExp algorithm (with $r=1$) guarantees a competitive ratio of $\frac{1}{8e}$ for any predictable reward-function. For the subsequent examples, we use the more restrictive \emph{binary rewards} setting, i.e., $V(i)\in\{0,1\}\fall i$; we also define $V^*=\{i\in N_I|V(i)=1\}$.

Now we show that given $\epsilon>0$, picking item $i$ for exploration with a probability proportional to $d_i^{-1\pm\epsilon}$ gives a \emph{vanishing competitive ratio}. Note that due to the symmetry of the graph, all user arrival patterns are equivalent. 
\begin{itemize}[nolistsep,noitemsep]
\item[$\bullet$] Consider an algorithm that picks items to explore with probability proportional to $d_i^{-1+\epsilon}$. Choose rewards such that $V^*=\{1,2,\ldots,n\}$, i.e., the first $n$ items. Since each of these items is connected to the user with same index, therefore $R^*(u)=1$. Now for user $1$, the probability of picking item $1$ is given by $p_{11}=\frac{1}{1+n\cdot n^{\epsilon-1}}=\frac{1}{1+n^{\epsilon}}
$, and for user $k+1$, irrespective of the choices made by the previous users, we can show that:
\begin{equation*}
p_{(k+1)(k+1)}\leq\frac{1}{(n-k)\cdot(n)^{\epsilon-1}}
\end{equation*}
Thus we have that the total reward summed across all users obeys:
\begin{equation*}
\EE[R_{alg}]\leq\sum_{k=1}^n\frac{1}{(n-k)\cdot(n)^{\epsilon-1}}=O(n^{1-\epsilon}\log n)
\end{equation*}
Finally, by symmetry we have that the per-user competitive ratio is the same as the ratio of sum reward to sum of optimal rewards for all users. Hence $\gamma(G,1)=O(n^{-\epsilon}\log n)=o(n)$.
\item[$\bullet$] Next consider an algorithm that picks items to explore with probability proportional to $d_i^{-1-\epsilon}$. Let $V^*=\{n+1\}$, i.e., the $(n+1)^{st}$ item, which is now connected to all users, and hence the sum of optimal rewards over all users is $R^*=n$. Now the probability that \emph{item $n+1$ is first explored by user $k+1$} is given by:
\begin{align*}
p_{k+1}&\leq\left(\frac{1}{1+(n-k)\cdot n^{-1-\epsilon}}\right)^k\left(\frac{n^{-1-\epsilon}}{1+(n-k)\cdot n^{-1-\epsilon}}\right)\\
&=\frac{n^{k+k\epsilon}}{(n-k+n^{1+\epsilon})^{k+1}}\leq n^{-1-\epsilon}
\end{align*}
Thus we have:
\begin{align*}
\EE[R_{alg}]&=\sum_{k=0}^{n-1}p_{k+1}\cdot(n-k)\leq\sum_{j=1}^{n}jn^{-1-\epsilon}=O(n^{1-\epsilon})
\end{align*}
and hence (again via symmetry arguments) $\gamma(G,1)=O(n^{-\epsilon})=o(n)$.
\end{itemize}
\Halmos
\endproof

\subsection{Negative Result: Deterministic Policies For Explore Vs. Exploit}

\proof{Proof of Theorem \ref{thm:negexploit}.}

\noindent\emph{Exploit-when-possible is not competitive:} In Exploit-when-possible, a user $u$ exploits whenever a non-zero valued item is available in $\mathcal{N}(u)\cap N_I^{expl}$, and explores otherwise (via some arbitrary policy). Given any $\epsilon\in(0,1)$, we consider the complete bipartite graph on $n\times n$ nodes (i.e., $n_I=n_U=n$). We consider the item values to be generated as follows: an item $i^*\in N_I$ is picked uniformly at random, and $V$ is defined to be:
\begin{equation*}
V(i)=\begin{dcases}
   1 & :i=i^* \\
   \delta & :i\neq i^* 
  \end{dcases}
\end{equation*}

Again, via symmetry arguments, we focus on the ratio of the sum of rewards earned by the algorithm to the sum of optimal rewards across users. For an \emph{arbitrary} user-arrival pattern $\pi$, we have that an algorithm which exploits whenever a user has a neighboring item with value $>0$ will earn an expected reward of $R_1^{alg}=1+n\delta(1-\frac{1}{n})$, 
and hence $\gamma_1^{alg}=\delta+\frac{1-\delta}{n}$. Now given $\epsilon>0$, we can choose $n\geq\frac{1}{\epsilon}$ and $\delta<\frac{n\epsilon-1}{n-1}$ to get $\gamma_1^{alg}<\epsilon$.

\noindent\emph{Exploit-above-fixed-threshold is not competitive:} A possible fix for the above problem could be to assume that a user $u$ exploits whenever a the highest-valued item in $\mathcal{N}(u)\cap N_I^{expl}$ has a value greater than some threshold $t\in\setR_+$ (and explores otherwise, again via some arbitrary policy); note that such a deterministic threshold is part of the algorithm specification and hence assumed to be known to the adversary. Similar to the previous case, we now show that such a strategy is also non-competitive.

Consider again the complete bipartite graph on $n\times n$ nodes (i.e., $n_I=n_U=n$). Given a threshold $t\in\setR_+$, we pick an item $i^*\in N_I$ uniformly at random and define $V$ as:
\begin{equation*}
V(i)=\begin{dcases}
   \delta & :i=i^* \\
   0 & :i\neq i^* 
  \end{dcases}
\end{equation*}
Then clearly $R^*_1=n\delta$. Now suppose $\delta<t$; for any arbitrary user-arrival pattern $\pi$ we have that an algorithm which exploits above threshold $t$ will give an expected reward of $R_1^{alg}=\delta$ and hence $\gamma_1^{alg}=\frac{1}{n}$. Again, given $\epsilon>0$, we choose $n\geq\frac{1}{\epsilon}$ to get $\gamma_1^{alg}<\epsilon$.
\Halmos
\endproof

\section{Inferring Item-Values from Multiple Ratings}
\label{appsec:multi}

\proof{Proof of Theorem \ref{thm:fexplore}}
As before, for $k\in[r]$, we define $\mathds{1}^{\mbox{\tiny ULExp-$f$}}_{s\rightarrow I^*_k(s)}$ to be $1$ if the user corresponding to visit $s$ is presented with the $k^{th}$ highest valued available item. Suppose $\delta(f)=0$, i.e., an item's value is exactly known after $f$ explorations. Then, from Lemma \ref{lem:minind}, we have that for any visit $s$, $\EE\left[R_r(s)\right]\geq\EE\left[\min_{k\in[r]}\EE\left[\mathds{1}^{\mbox{\tiny ULExp-$f$}}_{s\rightarrow I^*_k(s)}\right]R^*_r(s)\right]$, where the inner expectation is over the randomness in the algorithm, and the outer expectation is over randomness in the sample path. Now since we assume that algorithms know the value of pre-explored items (defined now as those which have been explored at least $f$ times) have their value known to within a multiplicative factor of $(1\pm\delta(f))$, then we have that when $I^*_k(s)$ is in $N_I^{expl}$, then with probability $\frac{1}{f+1}$, either it is explored or a lower valued item is explored, with a value at least $(1\pm\delta(f))V(I^*_k(s))$. In the worst case, this affects all $r$ top items; via linearity of expectation, we get:
\begin{equation*}
\EE\left[R_r(s)\right]\geq(1-2\delta(f))\EE\left[\min_{k\in[r]}\EE\left[\mathds{1}^{\mbox{\tiny ULExp-$f$}}_{s\rightarrow I^*_k(s)}\right]R^*_r(s)\right]
\end{equation*}

To complete the proof, for any user-visit $s$, and for all $k\in[r]$, we need to show that the corresponding $k^{th}$-top item, denoted $I_k^*(s)$ satisfies $\EE\left[\mathds{1}^{\mbox{\tiny ULExp}}_{s\rightarrow i^*_k(s)}\right]\geq \frac{1}{(f+1)^{f+1}}\left(\frac{r}{(5Z_{\max}(G)+2)}\right)^f$. Note that for $I^*_k(s)$, visit $s$ may correspond to one the first $f$ neighboring users, or may come after $I^*_k(s)$ has already had $f$ chances to be viewed -- in the latter case, in order to be presented to $s$, we need that the item $I^*_k(s)$ have been seen by all of its first $f$ neighboring users. Since this is the least likely of the above events, we need to lower bound this probability to complete the proof.

In the proof of Theorem \ref{thm:ulexp2}, we had shown that for any visit $s$ and any $k\in[r]$, the latest-items set $L(I^*_k(s))$ satisfied $\EE[|L(I^*_k(s))|]\leq 5Z_{\max}+2$. Now for the same user, and for any of its first $f$ neighboring visiting-users, we have that under the ULExp algorithm, it is picked independently with probability at least $\frac{rf}{f+1}\cdot\frac{1}{f}\cdot\frac{1}{\EE[|(I^*_k(s))|]}$. Thus the probability that it is explored by \emph{all $f$} of its first visiting users is at least $\left(\frac{r}{(f+1)(5Z_{\max}+2)}\right)^f$. Furthermore, the user corresponding to $s$ exploits the $k^{th}$ highest-valued available item with probability at least $\frac{1}{f+1}$. Combining these, we get the result. 
\Halmos
\endproof

\end{APPENDICES}

\ACKNOWLEDGMENT{This work was supported by NSF Grants
  CNS-1017525 and CNS-1320175, and an Army Research Office Grant W911NF-11-1-0265.}

\end{document}